\documentclass[journal]{IEEEtran}

\usepackage{amsfonts}
\usepackage{balance}
\usepackage{booktabs}
\usepackage{amsmath}
\usepackage{tabularx}
\usepackage{threeparttable}
\usepackage{subfigure}
\usepackage{graphicx}
\usepackage{multirow}
\usepackage{rotating}
\usepackage{algorithmic}
\usepackage{color}
\usepackage{url}
\usepackage[linesnumbered,ruled]{algorithm2e}

\ifCLASSOPTIONcompsoc
  \usepackage[nocompress]{cite}
\else
  \usepackage{cite}
\fi

\ifCLASSINFOpdf
\else
\fi

\begin{document}

\title{Toward Multidiversified Ensemble Clustering of High-Dimensional Data:\\ From Subspaces to Metrics and Beyond}

\author{Dong Huang,~\IEEEmembership{Member,~IEEE, }
        Chang-Dong Wang,~\IEEEmembership{Member,~IEEE, }\\
        Jian-Huang Lai,~\IEEEmembership{Senior Member,~IEEE, }
        and~Chee-Keong Kwoh,~\IEEEmembership{Senior Member,~IEEE}
\IEEEcompsocitemizethanks{\IEEEcompsocthanksitem Dong Huang is with the College of Mathematics and Informatics, South China Agricultural University, Guangzhou, China, and also with Pazhou Lab, Guangzhou, China. E-mail: huangdonghere@gmail.com.
\IEEEcompsocthanksitem Chang-Dong Wang and Jian-Huang Lai are with the School of Computer Science and Engineering,
Sun Yat-sen University, Guangzhou, China, and also with Guangdong Key Laboratory of Information Security Technology, Guangzhou, China, and also with Key Laboratory of Machine Intelligence and Advanced Computing, Ministry of Education, China.
E-mail: changdongwang@hotmail.com, stsljh@mail.sysu.edu.cn.
\IEEEcompsocthanksitem Chee-Keong Kwoh is with the School of Computer Science and Engineering, Nanyang Technological University, Singapore. E-mail: asckkwoh@ntu.edu.sg.}
}

\maketitle

\begin{abstract}
The rapid emergence of high-dimensional data in various areas has brought new challenges to current ensemble clustering research. To deal with the curse of dimensionality, recently considerable efforts in ensemble clustering have been made by means of different subspace-based techniques. However, besides the emphasis on subspaces, rather limited attention has been paid to the potential diversity in similarity/dissimilarity metrics. It remains a surprisingly open problem in ensemble clustering how to create and aggregate a large population of diversified metrics, and furthermore, how to jointly investigate the multi-level diversity in the large populations of metrics, subspaces, and clusters in a unified framework. To tackle this problem, this paper proposes a novel multidiversified ensemble clustering approach. In particular,  we create a large number of diversified metrics by randomizing a scaled exponential similarity kernel, which are then coupled with random subspaces to form a large set of metric-subspace pairs. Based on the similarity matrices derived from these metric-subspace pairs, an ensemble of diversified base clusterings can thereby be constructed. Further, an entropy-based criterion is utilized to explore the cluster-wise diversity in ensembles, based on which three specific ensemble clustering algorithms are presented by incorporating three types of consensus functions. Extensive experiments are conducted on 30 high-dimensional datasets, including 18 cancer gene expression datasets and 12 image/speech datasets, which demonstrate the superiority of our algorithms over the state-of-the-art. The source code is available at https://github.com/huangdonghere/MDEC.
\end{abstract}


\IEEEpeerreviewmaketitle

\section{Introduction}\label{sec:introduction}

\IEEEPARstart{T}{he} last decade has witnessed significant progress in the development of the ensemble clustering technique \cite{Li_WCC08,iam_on11_linkbased,wang11_tsmcb,li12_nips,Zheng14_TKDD,Jing15_pr,wu15_TKDE,Huang16_TKDE,huang15_ecfg,Liu2017_DMKD,liu17_tkde,liu17_bioinformatics,yu18_tkde,shi18_tcyb,huang17_tcyb,huang18_tsmcs,huang19_tkde,bai18_tfs,bai19_tkde,tao19_tnnls,yu19_tkde,li19_ai}, which is typically featured by its ability of combining multiple base clusterings into a probably better and more robust consensus clustering and has recently shown promising advantages in discovering clusters of arbitrary shapes, dealing with noisy data, coping with data from multiple sources, and producing robust clustering results \cite{wu15_TKDE}.

In recent years, with high-dimensional data widely appearing in various areas, new challenges have been brought to the conventional ensemble clustering algorithms, which, however, often lack the ability to well address the high-dimensional issues. As is called the curse of dimensionality, it is highly desired but very difficult to find the inherent cluster structure hidden in the huge dimensions, especially when it is frequently coupled with quite low sample size. Recently some efforts have been devoted to ensemble clustering of high-dimensional data, which typically exploit some subspace-based (or feature-based) techniques (such as random subspace sampling \cite{Yu07_bioinformatics,yu15_tkde,Yu16_tkde_incremental,Yu17_tkde,yu18_tkde}, stratified subspace sampling \cite{Jing15_pr}, and subspace projection \cite{fern2003random}) to explore the diversity in high-dimensionality. Inherently, these subspace-based techniques select or linearly combine data features into different subsets (i.e., subspaces) by a variety of strategies to seek more perspectives for finding cluster structures.

Besides the issue of subspaces (or features), the choice of similarity/dissimilarity metrics is another crucial factor in dealing with high-dimensional data \cite{Lee14_TVCG,Hsu09}. The existing ensemble clustering methods generally adopt one or a few preselected metrics, which are often selected explicit or implicitly based on the expert's knowledge or some prior assumptions. However, few, if not none, of them have considered the potentially huge opportunities hidden in randomized metric spaces. On the one hand, it is very difficult to select or learn an optimal metric for a given dataset without human supervision or implicit assumptions. On the other hand, with different metrics capable of reflecting different perspectives on data, the joint use of a large number of randomized/diversified metrics may reveal more valuable information hidden in high-dimensionality. However, it is surprisingly still an open problem in ensemble clustering how to produce and aggregate a large number of diversified metrics to enhance the consensus performance. Furthermore, starting from the metric diversification problem, another crucial challenge arises as to how to jointly exploit multiple levels of diversity in the large number of metrics, subspaces, and clusters, in a unified ensemble clustering framework.

To tackle the above-mentioned problem, this paper presents a novel ensemble clustering approach termed multidiversified ensemble clustering (MDEC) by jointly exploiting large populations of diversified metrics, random subspaces, and weighted clusters. Particularly, a scaled exponential similarity kernel is utilized as the seed kernel, which has advantage in parameter flexibility and neighborhood adaptivity and is randomized to breed a large set of diversified metrics. Then, the set of diversified metrics are coupled with random subspaces to form a large number of metric-subspace pairs, which contribute to the jointly randomized ensemble generation process where the set of diversified base clusterings can thereby be produced with the help of  spectral clustering. Further, to exploit the cluster-wise diversity in the multiple base clusterings, an entropy-based cluster validity strategy is adopted to evaluate and weight each base cluster by considering the distribution of clusters in the entire ensemble, based on which three specific ensemble clustering algorithms are therefore proposed by incorporating three types of consensus functions. Extensive experiments are conducted on 30 real-world high-dimensional datasets, including 18 cancer gene expression datasets and 12 image/speech datasets. The experimental results have shown the superiority of our approach against the state-of-the-art ensemble clustering approaches for clustering high-dimensional data.

For clarity, the main contributions of this work are summarized as follows:
\begin{enumerate}
  \item This paper for the first time, to our knowledge, shows that the joint use of a large population of randomly diversified metrics can significantly benefit the ensemble clustering of high-dimensional data in a purely unsupervised manner.
  \item A new metric diversification strategy is proposed by randomizing the scaled exponential similarity kernel with both parameter flexibility and neighborhood adaptivity considered, which is further coupled with random subspace sampling for the jointly randomized generation of base clusterings.
  \item A new ensemble clustering framework termed MDEC is presented, which is capable of simultaneously exploiting large populations of diversified metrics, random subspaces, and weighted clusters in a unified model for high-dimensional data.
  \item Three specific algorithms are designed under the proposed MDEC framework by incorporating three types of consensus functions. Experiments conducted on a variety of real-world high-dimensional datasets have confirmed the advantages of the proposed algorithms over the state-of-the-art.
\end{enumerate}

The remainder of the paper is organized as follows. The related work is reviewed in Section~\ref{sec:related_work}. The proposed ensemble clustering framework is described in Section~\ref{sec:framework}. The experimental results are reported in Section~\ref{sec:experiment}. Finally, the paper is concluded in Section~\ref{sec:conclude}.

\section{Related Work}
\label{sec:related_work}

Due to its ability of combining multiple base clusterings into a probably better and more robust consensus clustering, the ensemble clustering technique has been receiving increasing attention in recent years. Many ensemble clustering algorithms have been developed from different technical perspectives \cite{iam_on11_linkbased,wu15_TKDE,huang15_ecfg,Huang16_TKDE,huang17_tcyb,Liu2017_DMKD,liu17_tkde,liu17_bioinformatics,Fred05_EAC,Li_WCC08,Zheng14_TKDD,wang11_tsmcb,li12_nips,wu13_ijcai,Zhong15_pr,huang14_weac,Fan17_ijcai,Yousefnezhad17_tcyb,Huang16_neucom,wang09_pr,yi_icdm12,strehl02,fern04_bipartite,topchy05,franek13_pr}, which can be classified into three main categories, namely, the pair-wise co-occurrence based methods, the graph partitioning based methods, and the median partition based methods.

The pair-wise co-occurrence based methods \cite{Fred05_EAC,wang09_pr,yi_icdm12} typically construct a co-association matrix by considering the frequency that two data samples occur in the same cluster among the multiple base clusterings. The co-association matrix is then used as the similarity matrix for the data samples, upon which some clustering algorithms can thereby be performed to obtain the final clustering result. Fred and Jain \cite{Fred05_EAC} first introduced the concept of the co-association matrix and proposed the evidence accumulation clustering (EAC) method, which applied a hierarchical agglomerative clustering algorithm \cite{jain10_survey} on the co-association matrix to build the consensus clustering. To extend the EAC method, Wang et al. \cite{wang09_pr} took the cluster sizes into consideration and proposed the probability accumulation method. Yi et al. \cite{yi_icdm12} dealt with the uncertain entries in the co-association matrix by first labeling them as unobserved, and then recovering the unobserved entries by the matrix completion technique. Liu et al. \cite{liu17_tkde} proved that the spectral clustering of the co-association matrix is equivalent to a weighted version of $K$-means, and proposed the spectral ensemble clustering (SEC) method to effectively and efficiently obtain the consensus result.

The graph partitioning based methods \cite{Huang16_TKDE,strehl02,fern04_bipartite} generally construct a graph model for the ensemble of multiple base clusterings, and then partition the graph into several disjoint subsets to obtain the final clustering result. Strehl and Ghosh \cite{strehl02} solved the ensemble clustering problem by using three graph partitioning based algorithms, namely, cluster-based similarity partitioning algorithm (CSPA), hypergraph partitioning algorithm (HGPA), and meta-clustering algorithm (MCLA). Fern and Brodley \cite{fern04_bipartite} formulated a bipartite graph model by treating both clusters and data samples as nodes, and partitioned the graph by the METIS algorithm \cite{karypis98_METIS} to obtain the consensus result. Huang et al. \cite{Huang16_TKDE} dealt with the ensemble clustering problem by sparse graph representation and random walk trajectory analysis, and presented the probability trajectory based graph partitioning (PTGP) method.

The median partition based methods \cite{huang15_ecfg,topchy05,franek13_pr} typically formulate the ensemble clustering problem into an optimization problem which aims to find the median partition such that the similarity between the base partitions (i.e., base clusterings) and the median partition is maximized. The median partition problem is NP-hard \cite{topchy05}. To find an approximate solution, Topchy et al. \cite{topchy05} cast the median partition problem into a maximum likelihood problem and solved it by the EM algorithm. Franek and Jiang \cite{franek13_pr} reduced the ensemble clustering problem to an Euclidean median problem and solved it by the Weiszfeld algorithm \cite{Weiszfeld09}. Huang et al. \cite{huang15_ecfg} formulated the ensemble clustering problem into a binary linear programming problem and obtained an approximate solution based on the factor graph model and the max-product belief propagation \cite{Kschischang_FG_SPA:01}.

Although in recent years significant advances have been made in the research of ensemble clustering \cite{iam_on11_linkbased,wu15_TKDE,huang15_ecfg,Huang16_TKDE,huang17_tcyb,Liu2017_DMKD,liu17_tkde,liu17_bioinformatics,Fred05_EAC,Li_WCC08,wang11_tsmcb,li12_nips,Zheng14_TKDD,wu13_ijcai,Zhong15_pr,huang14_weac,Fan17_ijcai,Yousefnezhad17_tcyb,Huang16_neucom,wang09_pr,yi_icdm12,strehl02,fern04_bipartite,topchy05,franek13_pr}, yet the existing methods are mostly devised for general-purpose scenarios and lack the desirable ability to appropriately address the clustering problem of high-dimensional data. More recently, some efforts have been made to deal with the curse of dimensionality, where subspace-based (or feature-based) techniques are often exploited. Jing et al. \cite{Jing15_pr} adopted stratified feature sampling to generate a set of subspaces, which are further incorporated into several ensemble clustering algorithms to build the consensus clustering for high-dimensional data. Yu et al. \cite{yu15_tkde} proposed a novel subspace-based ensemble clustering framework termed AP$^2$CE, which integrates random subspaces, affinity propagation, normalized cut, and five candidate distance metrics. Further, Yu et al. \cite{Yu16_tkde_incremental} proposed a semi-supervised subspace-based ensemble clustering framework by incorporating random subspaces, constraint propagation, incremental ensemble selection, and normalized cut into the framework. Fern and Brodley \cite{fern2003random} exploited random subspace projection to build a set of subspaces, which in fact are obtained by (randomly) linear combination of features (or feature sets). Chu et al. \cite{chu20_tkde} used three clustering results (generated by three different clustering algorithms) to guide the feature learning process. These methods \cite{chu20_tkde,Jing15_pr,yu15_tkde,Yu16_tkde_incremental,fern2003random} typically exploit the diversity in high-dimensionality by various subspace-based or feature learning based techniques, but few of them have fully considered the potentially huge diversity in metric spaces. The existing methods \cite{chu20_tkde,Jing15_pr,yu15_tkde,Yu16_tkde_incremental,fern2003random} generally use one or a few preselected similarity/disimilarity metrics, which are selected implicitly based on the expert's knowledge or some prior assumptions. Although the method in \cite{yu15_tkde} proposed to randomly select a metric out of the five candidate metrics at each time, yet it still failed to go beyond a few metrics to explore the huge potential hidden in a large number of diversified metrics, which may play a crucial role in clustering high-dimensional data. The key challenge here lies in how to create such a large number of highly diversified metrics, and further how to jointly exploit the diversity in the large number of metrics, together with subspace-wise diversity and cluster-wise diversity, to achieve a unified ensemble clustering framework for high-dimensional data.

\section{Proposed Framework}
\label{sec:framework}

This section describes the overall algorithm of the proposed ensemble clustering framework. A brief overview is provided in Section~\ref{sec:overview}. The metric diversification process is presented in Section~\ref{sec:metric_diversification}. The jointly randomized ensemble generation is introduced in Section~\ref{sec:ensemble_generation}. Finally the consensus functions are given in Section~\ref{sec:consensus_function}.

\subsection{Brief Overview}
\label{sec:overview}
\begin{figure}[!t]
\begin{center}
{
{\includegraphics[width=0.92\columnwidth]{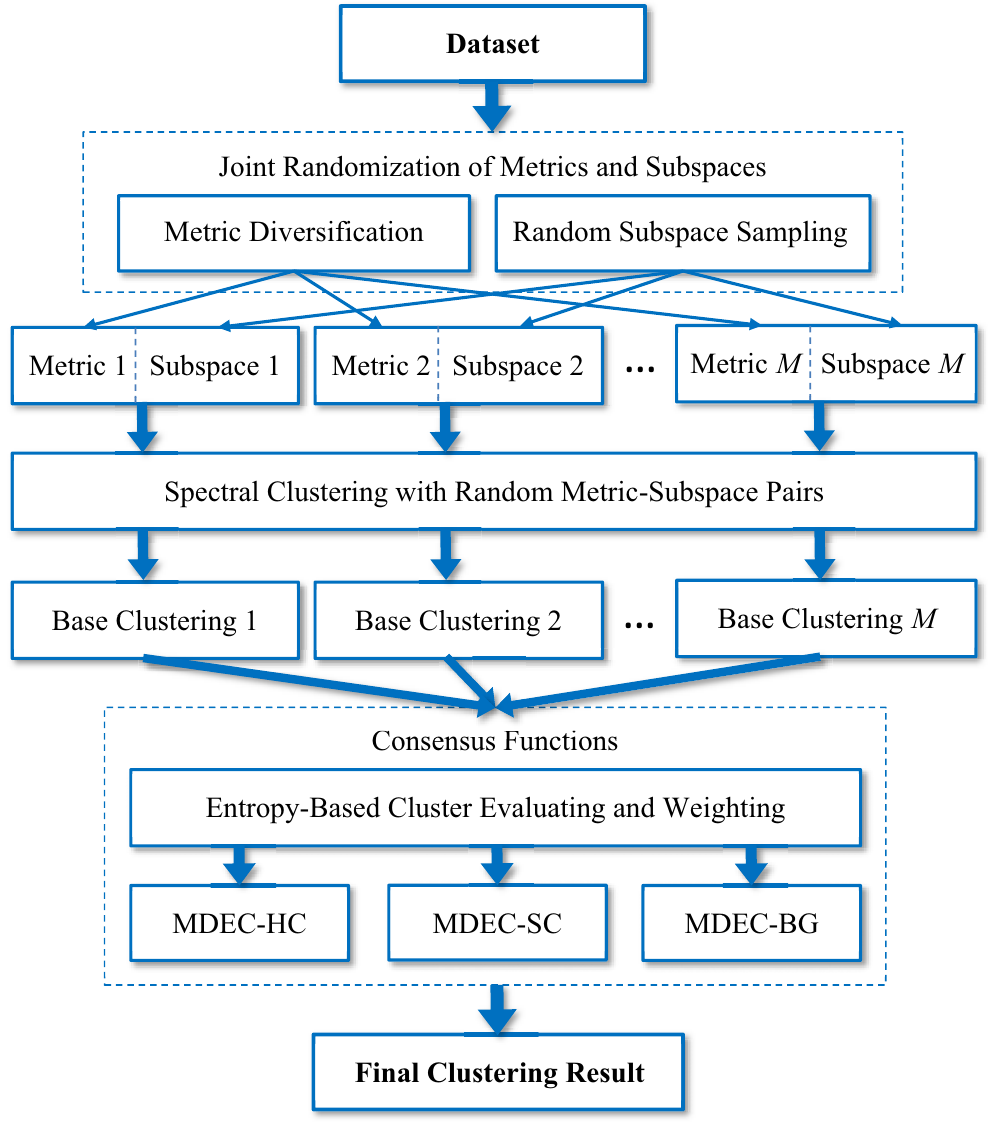}}}
\caption{Flow diagram of the proposed MDEC approach.}
\label{fig:flowchart}
\end{center}
\end{figure}

In this paper, we propose a novel multidiversified ensemble clustering (MDEC) framework (as shown in Fig.~\ref{fig:flowchart}). First, we create a large number of diversified metrics by randomizing a scaled exponential similarity kernel, and combine the diversified metrics with the random subspaces to form a large set of random metric-subspace pairs. Second, with each random metric-subspace pair, we construct a similarity matrix for the data samples. The spectral clustering is then performed on these similarity matrices derived from metric-subspace pairs to obtain an ensemble of base clusterings. Third, to exploit the cluster-wise diversity in the ensemble of multiple base clusterings, we adopt an entropy based criterion to evaluate and weight the clusters by considering the distribution of cluster labels in the entire ensemble. With the weighted clusters, a locally weighted co-association matrix is further constructed to serve as a summary of the ensemble. Finally, to obtain the consensus clustering, three types of consensus functions are presented based on hierarchical clustering (HC), spectral clustering (SC), and bipartite graph (BG) model, respectively, leading to three specific ensemble clustering algorithms termed MDEC-HC, MDEC-SC, and MDEC-BG, respectively. It is noteworthy that we simultaneously incorporate three levels of diversity, i.e., metric-wise diversity, subspace-wise diversity, and cluster-wise diversity, in a unified framework, which has shown its advantages in dealing with high-dimensional data when compared to the state-of-the-art ensemble clustering algorithms. In the following sections, we will further introduce each component of the proposed framework.

\subsection{Diversification of Metrics}
\label{sec:metric_diversification}

The choice of similarity/dissimilarity metrics plays a crucial role in the field of data mining and machine learning \cite{Wang13_pcyuen,xiong16_tkdd,li18_ijcv,liu19_tpami,chen19_tcyb,han19_nips}, especially for high-dimensional data analysis where the high-dimensionality further complicates the use of metrics \cite{xiong16_tkdd,han19_nips}. Some supervised or semi-supervised learning techniques have been developed to learn a suitable metric for some specific applications \cite{Wang13_pcyuen,xiong16_tkdd,li18_ijcv,liu19_tpami,chen19_tcyb,han19_nips}, but in unsupervised scenarios it is still very difficult to discover one or a few proper metrics given a task without prior knowledge.

Instead of relying on one or a few selected or learned metrics, this paper proposes to jointly use a large number of randomly diversified metrics, which will be further coupled with a large number of random subspaces, in a unified ensemble clustering framework for high-dimensional data. Here, two sub-problems are involved, namely, how to create a large number of diversified metrics, and how to collectively exploit them in ensemble clustering.

To create the diversified metrics, we take advantage of the kernel trick with randomization incorporated. The kernel similarity metrics have been proved to be a powerful tool for clustering complex data \cite{Wang13_pcyuen,liu19_tpami}, which, however, suffer from the difficulties in selecting proper kernel parameters. The kernel parameters can be learned by some metric learning techniques \cite{Wang13_pcyuen,liu19_tpami} with supervision or semi-supervision. But without human supervision, it is often extremely difficult to decide proper kernel parameters. This is a critical disadvantage of kernel methods for conventional (unsupervised) applications, which, nevertheless, just becomes an important advantage in our situation where what is highly desired is not the selection of a good kernel similarity metric, but the flexibility to create a large number of diversified ones.

Specifically, in this paper, we adopt a scaled exponential similarity kernel \cite{wang14_nmeth} as the seed kernel, which has advantage in parameter flexibility and neighborhood adaptivity and is then randomized to breed a large population of diversified metrics. Given a set of $N$ data samples $\mathcal{X}=\{x_1, \cdots, x_N\}$, where $x_i\in\mathbb{R}^D$ is the $i$-th sample and $D$ is the number of features. The kernel similarity between samples $x_i$ and $x_j$ is defined as:
\begin{align}
\varphi^{\mathcal{X}}(x_i, x_j)=\begin{cases}\exp{\left(-\frac{d(x_i,x_j)}{\mu\varepsilon_{ij}}\right)}, &\text{if~} x_i\in N_k(x_j)\\
&\text{or~}x_j\in N_k(x_i),\\
0, &\text{otherwise,}
\end{cases}
\end{align}
where $\mu$ is a hyperparameter, $\varepsilon_{ij}$ is a scaling term, $d(x_i,x_j)$ is the Euclidean distance between $x_i$ and $x_j$, and $N_k(x_i)$ denotes the set of the $k$ nearest neighbors of $x_i$. The average distance between $x_i$ and its $k$ nearest neighbors can be computed as
\begin{align}
\rho_k(x_i)=\frac{1}{k}\cdot\sum_{x_p\in N_k(x_i)}d\left(x_i, x_p\right).
\end{align}

Then, as suggested in \cite{wang14_nmeth}, to simultaneously take into consideration the neighborhood of $x_i$, the neighborhood of $x_j$, and their distance, the scaling term $\varepsilon_{ij}$ is defined as the average of $\rho_k(x_i)$, $\rho_k(x_j)$, and $d(x_i,x_j)$. That is
\begin{align}
\varepsilon_{ij}=\frac{\rho_k(x_i)+\rho_k(x_j)+d(x_i,x_j)}{3}.
\end{align}

The above kernel is a variant of the Gaussian kernel. It has two free parameters, i.e., the hyperparameter $\mu$ and the number of nearest neighbors $k$. The motivation to adopt the SES kernel as the seed kernel in our approach is two-fold. First, with the influence of the scaling term where the $k$-nearest-neighbors' information is incorporated, the kernel has the adaptivity to the neighborhood structure among data samples. Moreover, with each $k$ value corresponding to a specific neighborhood size, by randomizing the parameter $k$, multi-scale neighborhood information can be explored to enhance the diversity. Second, the two free parameters $k$ and $\mu$ in the kernel provide high flexibility for adjusting the influence of the kernel and can contribute to the desired diversity of the generated metrics by randomly perturbing the two parameters.

Specifically, we randomly select the two parameters $\mu\in[\mu_{min},\mu_{max}]$ and $k\in [k_{min}, k_{max}]$, respectively, as follows:
\begin{align}
\mu = \mu_{min} + \sigma_1(\mu_{max}-\mu_{min}),\label{eq:randomMu}\\
k = k_{min} + \lfloor\sigma_2(k_{max}-k_{min})\rfloor,\label{eq:randomk}
\end{align}
where $\sigma_1\in [0,1]$ and $\sigma_2\in [0,1]$ are two uniform random variables, and $\lfloor\space\space\rfloor$ outputs the floor of a real number.

Note that our objective is not to find a good pair of parameters $\mu$ and $k$, but to randomize them to yield a large population of diversified metrics. The parameters $\mu$ and $k$ are suggested to be randomly selected in a wide range to enhance the diversity.

By performing the random selection $M$ times, a set of $M$ pairs of $\mu$ and $k$ are obtained, which correspond to $M$ randomized kernel similarity metrics for the dataset $\mathcal{X}$, denoted as
\begin{align}
\varphi^{\mathcal{X}}_{\mu_1,k_1}(\cdot,\cdot), \varphi^{\mathcal{X}}_{\mu_2,k_2}(\cdot,\cdot), \cdots, \varphi^{\mathcal{X}}_{\mu_M,k_M}(\cdot,\cdot),
\end{align}
where $\mu_i$ and $k_i$ are the $i$-th pair of randomized parameters.

\subsection{Ensemble Generation by Joint Randomization}
\label{sec:ensemble_generation}

In high-dimensional data, the cluster structures are often hidden in different low-dimensional subspaces \cite{xiong16_tkdd,han19_nips}. Besides the subspaces, different metrics may also provide complementary information for investigating the high-dimensional space. In this section, with the set of diversified metrics generated, we proceed to couple the large number of diversified metrics with random subspaces to explore the rich information in various subspaces in high-dimensional data.

Specifically, let $\mathcal{F} = \{f_1, \cdots, f_D\}$ be the set of features in the dataset $\mathcal{X}$, where $f_i$ denotes the $i$-th feature. A random subspace is a set of a certain number of features that are randomly sampled from the original feature set. The cluster structure of high-dimensional data may be hidden in differen feature subspaces as well as in different metric spaces. In this paper, we propose to jointly exploit large populations of diversified metrics and random subspaces. Specifically, we perform random subspace sampling $M$ times to obtain $M$ random subspaces, denoted as $\mathcal{F}_1,\cdots,\mathcal{F}_M$, which lead to $M$ component datasets, denoted as $\mathcal{X}_1, \cdots, \mathcal{X}_M$. Note that each component dataset $\mathcal{X}_i$ has the same number of data samples as the original dataset $\mathcal{X}$, but its feature set $\mathcal{F}_i$ only consists of $d\leq D$ attributes that are randomly sampled from $\mathcal{F}$ with a sampling ratio $\tau \in(0,1]$. Obviously, if $\tau = 1$, then it means every subspace is in fact the original feature space, i.e., no sub-sampling actually happens. Here, with the random subspaces generated, we can couple each of them with a randomly diversified metric (as describe in Section~\ref{sec:metric_diversification}), and thus obtain $M$ random metric-subspace pairs, denoted as
\begin{align}
\varphi^{\mathcal{X}_1}_{\mu_1,k_1}(\cdot,\cdot), \varphi^{\mathcal{X}_2}_{\mu_2,k_2}(\cdot,\cdot), \cdots, \varphi^{\mathcal{X}_M}_{\mu_M,k_M}(\cdot,\cdot).
\end{align}

In terms of the $m$-th metric-subspace pair $\varphi^{\mathcal{X}_m}_{\mu_m,k_m}(\cdot,\cdot)$, the similarity between samples $x_i$ and $x_j$ is computed by first mapping $x_i$ and $x_j$ onto the subspace associated with the component dataset $\mathcal{X}_m$ and then computing their kernel similarity with the randomly selected parameters $\mu_m$ and $k_m$. Thus, we can obtain $M$ similarity matrices in terms of the $M$ metric-subspace pairs as follows:
\begin{align}
\mathcal{S} = \{S^{(1)}, S^{(2)}, \cdots, S^{(M)}\},
\end{align}
where the $m$-th similarity matrix (i.e., $S^{(m)}$) is constructed w.r.t. the $m$-th metric-subspace pair $\varphi^{\mathcal{X}_m}_{\mu_m,k_m}(\cdot,\cdot)$, denoted as
\begin{align}
S^{(m)}=\{S^{(m)}_{ij}\}_{N\times N},
\end{align}
where
\begin{align}
S^{(m)}_{ij} = \varphi^{\mathcal{X}_m}_{\mu_m,k_m}(x_i,x_j).
\end{align}
is the $(i,j)$-th entry in $S^{(m)}$. Obviously, according to the definition of the kernel, it holds that $S^{(m)}_{ij}\in(0,1]$ for any $x_i, x_j \in\mathcal{X}$. If samples $x_i$ and $x_j$ have the same feature values in the subspace associated with $\mathcal{X}_m$, then their similarity $S^{(m)}_{ij}$ reaches its maximum $1$.

Having constructed $M$ similarity matrices with diversified metric-subspace pairs, we then exploit the spectral clustering \cite{vonLuxburg2007} to construct the ensemble of base clusterings.
Specifically, for the $m$-th similarity matrix $S^{(m)}$, we treat each data sample as a graph node and build a similarity graph as follows:
\begin{equation}
G^{(m)}=(V,E^{(m)}),
\end{equation}
where $V=\mathcal{X}$ is the node set, and $E^{(m)}$ is the edge set. The edge weights are decided by the similarity matrix $S^{(m)}$, i.e., for any $x_i,x_j\in\mathcal{X}$, we have $E^{(m)}_{ij}=S^{(m)}_{ij}$. Let $K_m$ denote the number of clusters in the $m$-base clustering. The objective of spectral clustering is to partition the graph $G^{(m)}$ into $K_m$ disjoint subsets. To this end, we construct the normalized graph Laplacian $L_m$ as follows:
\begin{align}
L_m=I-D_m^{-1/2}S^{(m)}D_m^{-1/2},
\end{align}
where the degree matrix $D_m\in\mathbb{R}^{N\times N}$ is a diagonal matrix with its $(i,i)$-th entry defined as the sum of the $i$-th row of $S^{(m)}$. The eigenvectors corresponding to the first $K_m$ eigenvalues of $L_m$ are computed and then stacked to form a new matrix $U_m\in\mathbb{R}^{N\times K_m}$, where the $i$-th column of $U_m$ is the eigenvector corresponding to the $i$-th eigenvalue of $L_m$. Thereafter, the matrix $T_m\in\mathbb{R}^{N\times K_m}$ can be obtained from $U_m$ by normalizing the rows to norm 1.
By treating each row of $T_m$ as a data point in $\mathbb{R}^{K_m}$, we can cluster the rows into $K_m$ clusters by $K$-means discretization, and thereby obtain the $m$-th base clustering based on the similarity matrix $S^{(m)}$. Formally, the $m$-th base clustering is denoted as
\begin{align}
\pi^m = \{C^m_1, C^m_2,\cdots, C^m_{K_m}\},
\end{align}
where $C^m_i$ is the $i$-th cluster in $\pi^m$. It is obvious that the $K_m$ clusters in a base clustering cover the entire dataset, i.e., $\bigcup_{i=1}^{K_m}C^m_i=\mathcal{X}$, and two clusters in the same base clustering will not overlap with each other, i.e., $\forall i\neq j, C^m_i\bigcap C^m_j=\emptyset$.

Finally, based on the $M$ diversified similarity matrices, we can construct an ensemble of $M$ base clusterings, denoted as
\begin{align}
\Pi = \{\pi^1, \pi^2, \cdots, \pi^M\},
\end{align}
where $\pi^m$ is the $m$-th base clustering in the ensemble $\Pi$.

\subsection{Consensus Functions}
\label{sec:consensus_function}

This paper aims to integrate multiple levels of diversity in a unified ensemble clustering framework for high-dimensional data. With the metric-wise and subspace-wise diversity exploited in the jointly randomized ensemble generation, in this section, we proceed to explore the cluster-wise diversity in ensembles and incorporate three types of consensus functions to combine the multiple base clusterings into the final consensus clustering result.

With each base clustering consisting of a certain number of clusters, the entire ensemble can also be viewed as a large set of clusters from different base clusterings. To exploit the different reliability of different clusters and incorporate the cluster-wise diversity in the consensus function, here we adopt a local weighting strategy \cite{huang17_tcyb} to evaluate and weight the base clusters by jointly considering the distribution of cluster labels in the entire ensemble using an entropic criterion. Formally, we denote the ensemble of clusters as
\begin{align}
\mathcal{C} = \{C_1, C_2, \cdots,C_{N_c}\},
\end{align}
where $C_i$ is the $i$-th cluster and $N_c$ is the total number of clusters in the ensemble $\Pi$. Note that $N_c = \sum_{m=1}^M K_m$.

Each cluster is a set of data samples. To estimate the uncertainty of different clusters, the concept of entropy is utilized here \cite{huang17_tcyb}. Given a cluster $C_i\in\mathcal{C}$ and a base clustering $\pi^m\in\Pi$, the uncertainty (or entropy) of $C_i$ w.r.t. $\pi^m$ can be computed as
\begin{align}
\label{eq:entropy_one}
H^{\pi^m}(C_i) = -\sum_{j=1}^{K_m}p(C_i,C_j^m)\log_2 p(C_i,C_j^m),
\end{align}
where
\begin{equation}
\label{eq:entropy_one2}
p(C_i,C_j^m)=\frac{|C_i\bigcap C_j^m|}{|C_i|},
\end{equation}
is the proportion of data samples in $C_i$ that also appear in $C_j^m$. It is obvious that $p(C_i,C_j^m)\in [0,1]$, which leads to $H^{\pi^m}(C_i)\in [0,+\infty)$. If and only if all data samples in $C_i$ also occur in the same cluster in $\pi^m$, the uncertainty of $C_i$ w.r.t. $\pi^m$ reaches its minimum 0.

With the general assumption that the set of base clusterings are independent of each other, we can obtain the uncertainty (or entropy) of $C_i$ w.r.t. the entire ensemble $\Pi$ as follows
\begin{align}
\label{eq:entropy_all}
H^{\Pi}(C_i) = \sum_{m=1}^M H^{\pi^m}(C_i).
\end{align}

Intuitively, higher uncertainty indicates lower reliability for a cluster, which implies that the ensemble of base clusterings tend to disagree with the cluster and accordingly a smaller weight can be associated with it \cite{huang17_tcyb}. In particular, we proceed to compute a reliability index from the above-mentioned uncertainty measure, and exploit it as a cluster weighting term in our consensus function. The experimental analysis about the efficacy of the cluster weighting term will also be provided in Section~\ref{sec:cmpComponent}. Specifically, the ensemble-driven cluster index (ECI) is computed as an indication for the reliability of each cluster in the ensemble, which is defined as follows:
\begin{equation}
\label{eq:ECI}
ECI(C_i)=\exp{\left(-\frac{H^{\Pi}(C_i)}{M}\right)}.
\end{equation}

Obviously, for any $\pi^m\in\Pi$, it holds that $H^{\pi^m}(C_i)\in [0,+\infty)$, then we have $H^{\Pi}(C_i)\in [0,+\infty)$ and thereby $ECI(C_i)\in (0,1]$. Note that a larger value of ECI is associated with a cluster of lower uncertainty (i.e., higher reliability). If and only if the data samples in $C_i$ appear in the same cluster in all of the $M$ base clusterings (i.e., all base clusterings agree that the data samples in $C_i$ should belong to the same cluster), the uncertainty of $C_i$ w.r.t. $\Pi$ reaches its minimum 0 and the ECI of $C_i$ reaches its maximum 1.

The ECI measure serves as a reliability index for different clusters in the ensemble $\Pi$. By using ECI as a cluster-weighting term, the locally weighted co-association matrix can be obtained as follow:
\begin{align}
\label{eq:lwca}
A=&\{A_{ij}\}_{N\times N},\\
A_{ij} =& \frac{1}{M}\cdot\sum_{m=1}^M w_i^m\cdot\delta^m_{ij},\label{eq:lwca_aij}\\
w_i^m =& ECI\left(Cls^m(x_i)\right),\label{eq:lwca_wim}\\
\delta^m_{ij}=&\begin{cases}1,&\text{if~}Cls^m(x_i)=Cls^m(x_j),\label{eq:lwca_delta2}\\
0,&\text{otherwise,}
\end{cases}
\end{align}
where $Cls^m(x_i)$ denotes the cluster in $\pi^m$ that $x_i$ belongs to. Note that $w_i^m$ is the cluster weighting term which weights each cluster according to its ECI value, while $\delta^m_{ij}$ is the pair-wise co-occurrence term that indicates whether two samples occur in the same cluster in a base clustering $\pi^m$.

With the cluster-wise diversity explored, we further exploit three types of consensus functions to obtain the consensus clustering result, and thereby present three specific ensemble clustering algorithms, termed MDEC-HC, MDEC-SC, and MDEC-BG, respectively, under our MDEC framework.

\subsubsection{MDEC-HC}

In MDEC-HC, a hierarchical clustering based consensus function is incorporated, which constructs a dendrogram (i.e., a hierarchical cluster tree) by iterative region merging. Specifically, given a dataset $\mathcal{X}$, the $N$ data samples in $\mathcal{X}$ are treated as the set of initial regions, denoted as
\begin{align}
\mathcal{R}^{(0)} =\{R^{(0)}_1,\cdots, R^{(0)}_{|\mathcal{R}^{(0)} |}\},
\end{align}
where $R^{(0)}_i=\{x_i\}$ is the $i$-th initial region and $|\mathcal{R}^(0) |=N$ is the number of initial regions. Here, the locally weighted co-association matrix $A$ in Eq.~(\ref{eq:lwca}) is used as the similarity matrix for the initial regions, denoted as $Z^{(0)}=A$, upon which the iterative region merging can then be performed. In each iteration, the two regions with the highest similarity will be merged into a new and larger region.

Let $\mathcal{R}^{(t)} =\{R^{(t)}_1,\cdots, R^{(t)}_{|\mathcal{R}^{(t)} |}\}$ denote the set of regions after the $t$-th iteration, whose similarity matrix can be updated according to the average-linkage criterion after the region merging process, that is
\begin{align}
Z^{(t)}=&\{Z^{(t)}_{ij}\}_{|\mathcal{R}^{(t)} |\times |\mathcal{R}^{(t)} |},\\
Z^{(t)}_{ij}=&\frac{1}{|R^{(t)}_i|\cdot |R^{(t)}_j|}\sum_{x_u\in R^{(t)}_i, x_v\in R^{(t)}_j}a_{uv},
\end{align}
where $|\mathcal{R}^{(t)} |$ denotes the number of regions in the region set $\mathcal{R}^{(t)}$, and $|R^{(t)}_i|$ denotes the number of data samples in the region $R^{(t)}_i$. Note that, in each iteration, the number of regions decrements by one. Obviously, it holds that $|\mathcal{R}^{(t)}| = N-t$, and the dendrogram will be completed after exactly $N-1$ iterations. Each level of the dendrogram corresponds to a clustering result with a certain number of clusters, and therefore the final consensus clustering can be obtained by specifying a level in the dendrogram.

\subsubsection{MDEC-SC}

In MDEC-SC, a spectral clustering based consensus function is exploited. We first construct a graph with the data samples treated as graph nodes and the locally weighted co-association matrix $A$ used as the adjacent matrix, denoted as:
\begin{equation}
G=(V,E),
\end{equation}
where $V=\mathcal{X}$ is the node set, and $E$ is the edge set. The  edge weight between nodes $x_i$ and $x_j$ is defined as $E_{ij}=A_{ij}$. The normalized graph Laplacian is computed as
\begin{align}
L = I-D^{-1/2}AD^{-1/2},
\end{align}
where $D\in\mathbb{R}^{N\times N}$ is the degree matrix. Let $K$ be the number of clusters that we aim to obtain. Then we compute the eigenvectors corresponding to the first $K$ eigenvalues of $L$, which are further stacked to form a new matrix $U\in\mathbb{R}^{N\times K}$ with each eigenvector being a column of it. Thereafter, the matrix $U$ will be row-normalized, upon which $K$-means discretization will be performed to obtain the final clustering result.

\subsubsection{MDEC-BG}

In MDEC-BG, a bipartite graph based consensus function is presented. Different from MDEC-SC, we build a bipartite graph with both data samples and base clusters treated as graph nodes, and then perform bipartite graph partitioning to obtain the consensus clustering. Specifically, the bipartite graph is defined as
\begin{align}
\tilde{G}=(U,V,\tilde{E}),
\end{align}
where $U\bigcup V$ is the node set (with $U=\mathcal{X}$ and $V=\mathcal{C}$), and $\tilde{E}$ is the edge set. An edge between two nodes exists if and only if one of them is a data sample and the other one is a base cluster that contains this data sample. Typically, given two nodes $u_i\in \mathcal{X}$ and $v_j\in \mathcal{C}$, their edge weight will be decided by two factors, i.e., their belonging-to relationship and the reliability of the connected cluster, which can be estimated by the ECI index. Formally, the edge weight between $u_i\in \mathcal{X}$ and $v_j\in \mathcal{C}$ is defined as
\begin{align}
\tilde{E}_{ij} = \begin{cases} ECI(v_j), &\text{if~} u_i\in v_j,\\
0, &\text{otherwise.}
\end{cases}
\end{align}
Note that the ECI value reflects the reliability of a cluster w.r.t. the entire ensemble of base clusterings (as shown in Eqs.~(\ref{eq:entropy_one}), (\ref{eq:entropy_one2}), (\ref{eq:entropy_all}), and (\ref{eq:ECI})). By taking advantage of ECI to define the edge weights in the bipartite graph $\tilde{G}$, the clusters with higher reliability can exhibit greater influence on the bipartite graph and thus play a bigger role in the consensus process.
Then, with the bipartite graph constructed, we further adopt the transfer cut \cite{CVPR12_Li} to efficiently partition the graph nodes into several subsets. By treating the data samples in each subset as a final cluster, the consensus clustering result can be obtained.

\section{Experiments}
\label{sec:experiment}

In this section, we conduct experiments on a variety of real-world high-dimensional datasets to compare the proposed three ensemble clustering algorithms. against nine state-of-the-art algorithms.

\subsection{Datasets and Evaluation Measures}

In our experiments, 30 real-world high-dimensional datasets are used, including 18 cancer gene expression datasets \cite{deSouto08_gds} and 12 image/speech datasets. Among the 12 speech/image datasets, ISOLET \cite{Bache+Lichman:2013} is a speech recognition dataset, while the other 11 datasets, including COIL20 \cite{nene96_coil20},
MSRA25 \cite{nie19_tcyb},
Binary Alphadigits \cite{SamRoweisHomepage},
YaleB32 \cite{nie19_tcyb},
Semeion \cite{Bache+Lichman:2013},
UMist \cite{graham98_UMist},
Multiple Features \cite{Bache+Lichman:2013},
Flowers17	 \cite{nilsback06_flower17},
MNIST \cite{SamRoweisHomepage},
USPS \cite{SamRoweisHomepage},
and Gisette \cite{Bache+Lichman:2013}, are image datasets.
To simplify the description, in the following, the 30 benchmark datasets will be abbreviated as \emph{DS-1} to \emph{DS-30}, respectively (as shown in Tables~\ref{table:datasets_gene} and \ref{table:datasets_image}).

To evaluate the quality of the clustering result, two widely used evaluation measures are adopted, namely, normalized mutual information (NMI) \cite{strehl02} and adjusted Rand index (ARI) \cite{vinh2010_ARI}. Note that greater values of NMI and ARI indicate better clustering results.

\subsection{Baseline Methods and Experimental Settings}

In our experiments, we compare the proposed ensemble clustering algorithms with nine baseline algorithms, i.e.,
stratified sampling based cluster-based similarity partitioning algorithm (SSCSPA) \cite{Jing15_pr}, stratified sampling based hypergraph partitioning algorithm (SSHGPA) \cite{Jing15_pr}, stratified sampling based meta-clustering algorithm (SSMCLA) \cite{Jing15_pr}, $K$-means based consensus clustering (KCC) \cite{wu15_TKDE}, probability trajectory based graph partitioning (PTGP) \cite{Huang16_TKDE}, spectral ensemble clustering (SEC) \cite{liu17_tkde}, entropy based consensus clustering (ECC) \cite{liu17_bioinformatics}, locally weighted graph partitioning (LWGP) \cite{huang17_tcyb}, and self-paced clustering ensemble (SPCE) \cite{zhou20_spce}.

For the baseline algorithms, the parameters will be set as suggested by their corresponding papers. For the proposed algorithms, to produce a set of diversified metrics, the two kernel parameters $\mu$ and $k$ are randomized in the ranges of $[0.2,0.8]$ and $[5,20]$, respectively. To generate the ensemble of base clusterings, the ensemble size $M=100$ and the sampling ratio $\tau=0.5$ are used. The number of clusters in each base clustering is randomly selected in the range of $[2, \sqrt{N}]$. Further, the performances of our algorithms w.r.t. different ensemble sizes $M$ and different sampling ratios $\tau$ will also be evaluated in Sections~\ref{sec:compEnSize} and \ref{sec:cmpComponent}, respectively.

\begin{table}[!t]
\centering
\caption{Description of the eighteen cancer gene expression datasets.}
\label{table:datasets_gene}
\begin{center}
\begin{tabular}{p{2.45cm}<{\centering}|p{0.9cm}<{\centering}|p{0.9cm}<{\centering}p{1.1cm}<{\centering}p{1cm}<{\centering}}
\toprule
Dataset         &Abbr.  &\#Sample     &Dimension      &\#Class\\
\midrule
Bhattacharjee-2001	&\emph{DS-1}    &203	&1,543	&5\\
Bredel-2005   &\emph{DS-2}    &50	&1,739	&3\\
Chowdary-2006	&\emph{DS-3}    &104	&182	&2\\
Garber-2001   &\emph{DS-4}    &66	&4,553	&4\\
Golub-1999-v1	&\emph{DS-5}    &72	&1,877	&2\\
Golub-1999-v2	&\emph{DS-6}    &72	&1,877	&3\\
Nutt-2003-v1	&\emph{DS-7}    &50	&1,377	&4\\
Pomeroy-2002-v2	&\emph{DS-8}    &42	&1,379	&5\\
Su-2001	&\emph{DS-9}    &174	&1,571	&10\\
West-2001   &\emph{DS-10}    &49	&1,198	&2\\
Yeoh-2002-v1   &\emph{DS-11}    &248	&2,526	&2\\
Yeoh-2002-v2   &\emph{DS-12}    &248	&2,526	&6\\
Alizadeh-2000-v2	&\emph{DS-13}    &62	&2,093	&3\\
Alizadeh-2000-v3   &\emph{DS-14}    &62	&2,093	&4\\
Armstrong-2002-v2   &\emph{DS-15}    &72	&2,194	&3\\
Ramaswamy-2001	&\emph{DS-16}    &190	&1,363	&14\\
Risinger-2003   &\emph{DS-17}    &42	&1,771	&4\\
Tomlins-2006-v1   &\emph{DS-18}    &104	&2,315	&5\\
\bottomrule
\end{tabular}
\end{center}
\end{table}

\begin{table}[!t]
\centering
\caption{Description of the twelve image/speech datasets.}
\label{table:datasets_image}
\begin{center}
\begin{tabular}{p{2.55cm}<{\centering}|p{0.8cm}<{\centering}|p{0.9cm}<{\centering}p{1.1cm}<{\centering}p{1cm}<{\centering}}
\toprule
Dataset         &Abbr.  &\#Sample     &Dimension      &\#Class\\
\midrule
ISOLET &\emph{DS-19}    &7,797   &617   &26\\
COIL20	&\emph{DS-20} &1,440	&1,024	&20\\
MSRA25 &\emph{DS-21}    &1,799   &256   &12\\
Binary Alphadigits	&\emph{DS-22} &1,404	&320	&36\\
YaleB32 &\emph{DS-23}    &2,414   &1,024   &38\\
Semeion &\emph{DS-24}    &1,593   &256   &10\\
UMist	&\emph{DS-25} &575	&10,304	&20\\
Multiple Features   &\emph{DS-26} &2,000    &649    &10\\
Flowers17	&\emph{DS-27} &1,360	&30,000	&17\\
MNIST &\emph{DS-28}    &5,000   &784   &10\\
Gisette   &\emph{DS-29}    &7,000   &5,000   &2\\
USPS  &\emph{DS-30}    &11,000   &256   &10\\
\bottomrule
\end{tabular}
\end{center}
\end{table}

\begin{table*}[!t]\scriptsize
\centering
\caption{Average performances (w.r.t. NMI) over 100 runs by different ensemble clustering algorithms.
On each dataset, the highest three scores are highlighted in \textbf{bold}, while the highest one in [\textbf{bold and brackets}].}
\label{table:compare_ce_nmi}
\begin{threeparttable}
\begin{tabular}{|m{0.67cm}<{\centering}|m{0.9cm}<{\centering}m{0.97cm}<{\centering}m{0.97cm}<{\centering}m{0.97cm}<{\centering}m{0.97cm}<{\centering}m{0.97cm}<{\centering}m{0.97cm}<{\centering}m{0.97cm}<{\centering}m{1cm}<{\centering}m{1.2cm}<{\centering}m{1.15cm}<{\centering}m{1.2cm}<{\centering}|}
\hline
Dataset &SSCSPA     &SSHGPA     &SSMCLA     &KCC     &PTGP  &SEC    &ECC    &LWGP   &SPCE  &MDEC-HC   &MDEC-SC   &MDEC-BG\\
\hline
\hline
\emph{DS-1}	&20.63$_{\pm2.11}$	&19.06$_{\pm2.15}$	&26.71$_{\pm2.95}$	&26.88$_{\pm6.93}$	&39.86$_{\pm1.55}$	&27.56$_{\pm3.04}$	&35.90$_{\pm1.23}$	&42.10$_{\pm4.82}$	&[\textbf{53.53}$_{\pm2.27}$]	&\textbf{52.41}$_{\pm1.68}$	&\textbf{51.72}$_{\pm1.25}$	&51.64$_{\pm1.11}$\\
\hline
\emph{DS-2}	&\textbf{40.53}$_{\pm2.79}$	&32.54$_{\pm7.33}$	&33.88$_{\pm2.94}$	&36.12$_{\pm2.79}$	&31.50$_{\pm3.62}$	&36.01$_{\pm3.67}$	&36.52$_{\pm5.94}$	&31.56$_{\pm3.28}$	&\textbf{40.64}$_{\pm5.27}$	&[\textbf{45.23}$_{\pm3.35}$]	&38.44$_{\pm3.19}$	&38.58$_{\pm3.02}$\\
\hline
\emph{DS-3}	&60.26$_{\pm4.47}$	&73.58$_{\pm4.94}$	&47.99$_{\pm17.36}$	&60.13$_{\pm3.26}$	&8.15$_{\pm0.00}$	&50.27$_{\pm11.41}$	&61.18$_{\pm1.25}$	&8.15$_{\pm0.00}$	&23.83$_{\pm18.80}$	&[\textbf{85.97}$_{\pm0.00}$]	&[\textbf{85.97}$_{\pm0.00}$]	&[\textbf{85.97}$_{\pm0.00}$]\\
\hline
\emph{DS-4}	&15.02$_{\pm0.84}$	&17.03$_{\pm1.87}$	&14.49$_{\pm2.40}$	&15.34$_{\pm4.08}$	&14.20$_{\pm2.49}$	&13.57$_{\pm2.67}$	&14.89$_{\pm3.15}$	&12.44$_{\pm2.59}$	&[\textbf{27.76}$_{\pm1.53}$]	&21.77$_{\pm0.82}$	&\textbf{22.38}$_{\pm4.05}$	&\textbf{22.86}$_{\pm0.79}$\\
\hline
\emph{DS-5}	&42.42$_{\pm5.77}$	&46.91$_{\pm6.64}$	&62.53$_{\pm9.31}$	&55.65$_{\pm3.73}$	&57.11$_{\pm6.67}$	&49.37$_{\pm6.05}$	&53.46$_{\pm2.11}$	&56.45$_{\pm2.69}$	&51.02$_{\pm0.60}$	&[\textbf{74.92}$_{\pm13.84}$]	&\textbf{73.75}$_{\pm7.77}$	&\textbf{71.79}$_{\pm11.81}$\\
\hline
\emph{DS-6}	&48.35$_{\pm3.59}$	&51.73$_{\pm7.02}$	&42.31$_{\pm15.19}$	&53.68$_{\pm8.42}$	&67.32$_{\pm5.32}$	&59.31$_{\pm6.68}$	&67.51$_{\pm4.18}$	&65.98$_{\pm3.90}$	&45.22$_{\pm2.62}$	&[\textbf{76.17}$_{\pm5.42}$]	&\textbf{75.14}$_{\pm3.33}$	&\textbf{76.04}$_{\pm2.86}$\\
\hline
\emph{DS-7}	&40.61$_{\pm3.50}$	&37.22$_{\pm6.00}$	&36.71$_{\pm3.90}$	&46.31$_{\pm3.48}$	&\textbf{48.40}$_{\pm0.98}$	&46.53$_{\pm2.67}$	&43.21$_{\pm1.58}$	&44.79$_{\pm0.60}$	&[\textbf{49.78}$_{\pm2.27}$]	&\textbf{48.26}$_{\pm0.82}$	&45.50$_{\pm3.89}$	&47.36$_{\pm1.04}$\\
\hline
\emph{DS-8}	&50.30$_{\pm4.17}$	&56.29$_{\pm5.27}$	&40.85$_{\pm12.12}$	&57.55$_{\pm5.32}$	&58.67$_{\pm2.73}$	&52.92$_{\pm8.03}$	&56.58$_{\pm3.09}$	&55.53$_{\pm2.78}$	&63.20$_{\pm2.72}$	&[\textbf{66.16}$_{\pm2.44}$]	&\textbf{65.53}$_{\pm2.77}$	&\textbf{64.80}$_{\pm1.14}$\\
\hline
\emph{DS-9}	&52.05$_{\pm1.40}$	&52.72$_{\pm3.49}$	&44.66$_{\pm10.09}$	&57.22$_{\pm2.78}$	&57.02$_{\pm3.24}$	&56.91$_{\pm4.53}$	&56.95$_{\pm2.76}$	&53.15$_{\pm2.26}$	&58.26$_{\pm7.31}$	&\textbf{66.57}$_{\pm2.27}$	&[\textbf{69.55}$_{\pm0.93}$]	&\textbf{69.41}$_{\pm1.32}$\\
\hline
\emph{DS-10}	&29.05$_{\pm4.68}$	&[\textbf{35.75}$_{\pm1.25}$]	&29.21$_{\pm4.49}$	&31.17$_{\pm3.09}$	&31.18$_{\pm1.33}$	&28.29$_{\pm4.29}$	&31.22$_{\pm0.00}$	&4.08$_{\pm0.00}$	&21.37$_{\pm5.62}$	&32.03$_{\pm3.92}$	&\textbf{32.45}$_{\pm2.21}$	&\textbf{34.29}$_{\pm2.37}$\\
\hline
\emph{DS-11}	&20.20$_{\pm0.30}$	&18.60$_{\pm1.37}$	&4.56$_{\pm4.57}$	&79.18$_{\pm11.23}$	&84.56$_{\pm2.01}$	&56.61$_{\pm35.73}$	&85.59$_{\pm6.47}$	&1.92$_{\pm10.32}$	&41.78$_{\pm22.53}$	&[\textbf{94.35}$_{\pm2.49}$]	&\textbf{92.91}$_{\pm2.50}$	&\textbf{93.64}$_{\pm2.13}$\\
\hline
\emph{DS-12}	&31.48$_{\pm2.51}$	&32.65$_{\pm4.85}$	&9.42$_{\pm5.32}$	&33.40$_{\pm1.31}$	&34.93$_{\pm4.53}$	&32.85$_{\pm2.87}$	&36.09$_{\pm3.73}$	&26.44$_{\pm1.05}$	&41.48$_{\pm1.17}$	&\textbf{45.98}$_{\pm5.39}$	&\textbf{50.81}$_{\pm5.34}$	&[\textbf{57.01}$_{\pm3.79}$]\\
\hline
\emph{DS-13}	&57.11$_{\pm1.02}$	&48.88$_{\pm3.43}$	&56.71$_{\pm2.28}$	&65.40$_{\pm18.11}$	&\textbf{97.56}$_{\pm7.89}$	&59.52$_{\pm13.51}$	&70.32$_{\pm20.54}$	&71.92$_{\pm3.87}$	&91.10$_{\pm9.48}$	&[\textbf{98.34}$_{\pm3.44}$]	&\textbf{94.86}$_{\pm4.31}$	&93.38$_{\pm3.74}$\\
\hline
\emph{DS-14}	&42.77$_{\pm3.38}$	&49.40$_{\pm2.65}$	&49.02$_{\pm3.82}$	&52.26$_{\pm4.44}$	&\textbf{63.97}$_{\pm2.07}$	&50.11$_{\pm9.60}$	&63.47$_{\pm2.07}$	&53.01$_{\pm4.60}$	&[\textbf{65.99}$_{\pm2.49}$]	&\textbf{65.28}$_{\pm1.46}$	&63.61$_{\pm1.63}$	&63.02$_{\pm1.42}$\\
\hline
\emph{DS-15}	&66.82$_{\pm2.16}$	&49.13$_{\pm6.83}$	&66.67$_{\pm3.09}$	&50.34$_{\pm5.45}$	&75.00$_{\pm4.99}$	&41.77$_{\pm7.09}$	&72.08$_{\pm5.32}$	&58.02$_{\pm9.73}$	&61.86$_{\pm8.91}$	&[\textbf{80.90}$_{\pm2.71}$]	&\textbf{80.42}$_{\pm2.11}$	&\textbf{78.56}$_{\pm2.16}$\\
\hline
\emph{DS-16}	&50.44$_{\pm1.65}$	&54.22$_{\pm1.67}$	&18.71$_{\pm4.94}$	&45.13$_{\pm2.79}$	&38.42$_{\pm2.16}$	&38.05$_{\pm5.40}$	&43.57$_{\pm4.85}$	&29.19$_{\pm0.86}$	&52.51$_{\pm1.15}$	&\textbf{65.56}$_{\pm1.53}$	&\textbf{66.60}$_{\pm1.18}$	&[\textbf{66.64}$_{\pm1.16}$]\\
\hline
\emph{DS-17}	&30.98$_{\pm5.61}$	&30.90$_{\pm4.37}$	&16.20$_{\pm9.87}$	&29.09$_{\pm3.41}$	&31.32$_{\pm2.60}$	&28.12$_{\pm3.68}$	&31.93$_{\pm3.47}$	&15.57$_{\pm0.79}$	&32.05$_{\pm1.04}$	&[\textbf{36.79}$_{\pm2.93}$]	&\textbf{32.77}$_{\pm1.44}$	&\textbf{33.13}$_{\pm1.17}$\\
\hline
\emph{DS-18}	&38.12$_{\pm1.90}$	&35.95$_{\pm3.24}$	&42.68$_{\pm3.19}$	&44.86$_{\pm5.88}$	&44.66$_{\pm1.78}$	&37.92$_{\pm5.01}$	&40.50$_{\pm1.83}$	&42.61$_{\pm2.06}$	&46.57$_{\pm2.01}$	&[\textbf{61.11}$_{\pm2.59}$]	&\textbf{53.30}$_{\pm2.24}$	&\textbf{54.30}$_{\pm2.04}$\\
\hline
\emph{DS-19}	&69.12$_{\pm1.52}$	&60.32$_{\pm1.56}$	&73.81$_{\pm1.00}$	&68.43$_{\pm1.04}$	&72.96$_{\pm0.70}$	&68.71$_{\pm1.25}$	&70.06$_{\pm0.88}$	&74.91$_{\pm0.53}$	&72.35$_{\pm1.86}$	&[\textbf{77.09}$_{\pm0.77}$]	&\textbf{76.41}$_{\pm0.29}$	&\textbf{76.48}$_{\pm0.50}$\\
\hline
\emph{DS-20}	&78.42$_{\pm1.17}$	&76.77$_{\pm2.20}$	&75.38$_{\pm4.97}$	&72.33$_{\pm1.98}$	&67.00$_{\pm2.65}$	&72.06$_{\pm3.01}$	&74.26$_{\pm1.37}$	&80.16$_{\pm1.10}$	&86.01$_{\pm1.28}$	&[\textbf{91.66}$_{\pm0.75}$]	&\textbf{90.15}$_{\pm1.06}$	&\textbf{91.25}$_{\pm0.78}$\\
\hline
\emph{DS-21}	&59.41$_{\pm1.50}$	&61.66$_{\pm2.84}$	&65.07$_{\pm2.07}$	&61.05$_{\pm1.61}$	&63.67$_{\pm1.97}$	&59.79$_{\pm3.59}$	&62.70$_{\pm2.39}$	&66.29$_{\pm1.59}$	&69.31$_{\pm1.56}$	&\textbf{70.37}$_{\pm1.05}$	&[\textbf{71.78}$_{\pm1.64}$]	&\textbf{71.03}$_{\pm1.28}$\\
\hline
\emph{DS-22}	&55.31$_{\pm0.97}$	&46.99$_{\pm1.35}$	&54.45$_{\pm1.35}$	&54.21$_{\pm0.73}$	&55.06$_{\pm0.57}$	&56.41$_{\pm0.60}$	&54.79$_{\pm0.59}$	&51.67$_{\pm0.87}$	&\textbf{60.57}$_{\pm0.65}$	&58.06$_{\pm1.31}$	&[\textbf{62.54}$_{\pm0.62}$]	&\textbf{62.33}$_{\pm0.62}$\\
\hline
\emph{DS-23}	&10.63$_{\pm0.31}$	&11.86$_{\pm0.52}$	&11.10$_{\pm0.50}$	&10.11$_{\pm0.40}$	&10.15$_{\pm0.40}$	&10.39$_{\pm0.45}$	&10.47$_{\pm0.34}$	&9.40$_{\pm0.33}$	&27.29$_{\pm1.23}$	&\textbf{34.12}$_{\pm1.36}$	&\textbf{35.40}$_{\pm0.98}$	&[\textbf{37.54}$_{\pm1.25}$]\\
\hline
\emph{DS-24}	&55.69$_{\pm3.29}$	&35.82$_{\pm2.07}$	&57.51$_{\pm2.33}$	&53.22$_{\pm1.77}$	&62.64$_{\pm0.69}$	&54.39$_{\pm3.28}$	&53.61$_{\pm1.74}$	&61.93$_{\pm3.57}$	&62.27$_{\pm3.03}$	&[\textbf{71.33}$_{\pm0.64}$]	&\textbf{67.85}$_{\pm1.41}$	&\textbf{68.51}$_{\pm1.25}$\\
\hline
\emph{DS-25}	&57.93$_{\pm1.22}$	&65.52$_{\pm2.53}$	&58.69$_{\pm3.61}$	&60.85$_{\pm1.74}$	&62.67$_{\pm1.11}$	&60.49$_{\pm1.39}$	&61.37$_{\pm1.20}$	&62.50$_{\pm1.53}$	&69.84$_{\pm1.34}$	&[\textbf{83.97}$_{\pm1.77}$]	&\textbf{82.24}$_{\pm2.73}$	&\textbf{83.15}$_{\pm2.45}$\\
\hline
\emph{DS-26}	&81.34$_{\pm3.10}$	&69.94$_{\pm3.07}$	&86.33$_{\pm2.11}$	&73.08$_{\pm3.97}$	&83.45$_{\pm2.14}$	&64.14$_{\pm6.06}$	&75.24$_{\pm1.57}$	&87.02$_{\pm1.64}$	&82.00$_{\pm4.02}$	&\textbf{90.68}$_{\pm2.09}$	&[\textbf{94.19}$_{\pm0.30}$]	&\textbf{93.99}$_{\pm0.26}$\\
\hline
\emph{DS-27}	&21.75$_{\pm0.75}$	&18.75$_{\pm1.24}$	&23.74$_{\pm0.59}$	&24.15$_{\pm0.47}$	&24.96$_{\pm0.39}$	&24.18$_{\pm0.49}$	&24.16$_{\pm0.39}$	&21.60$_{\pm0.75}$	&[\textbf{37.09}$_{\pm0.38}$]	&25.44$_{\pm0.64}$	&\textbf{27.56}$_{\pm0.39}$	&\textbf{27.14}$_{\pm0.38}$\\
\hline
\emph{DS-28}	&53.13$_{\pm3.25}$	&38.54$_{\pm2.66}$	&58.34$_{\pm2.20}$	&53.46$_{\pm3.18}$	&64.53$_{\pm1.64}$	&53.10$_{\pm2.26}$	&53.42$_{\pm2.23}$	&60.13$_{\pm1.22}$	&51.77$_{\pm8.95}$	&[\textbf{78.47}$_{\pm2.19}$]	&\textbf{76.04}$_{\pm1.20}$	&\textbf{77.92}$_{\pm0.96}$\\
\hline
\emph{DS-29}	&20.44$_{\pm6.59}$	&19.28$_{\pm10.95}$	&47.98$_{\pm4.69}$	&41.41$_{\pm9.87}$	&49.82$_{\pm0.87}$	&10.73$_{\pm6.22}$	&46.07$_{\pm2.01}$	&5.62$_{\pm14.21}$	&9.23$_{\pm0.54}$	&\textbf{67.23}$_{\pm5.28}$	&\textbf{67.43}$_{\pm0.68}$	&[\textbf{67.94}$_{\pm0.34}$]\\
\hline
\emph{DS-30}	&47.98$_{\pm2.88}$	&34.33$_{\pm2.08}$	&53.71$_{\pm2.89}$	&51.17$_{\pm1.52}$	&65.00$_{\pm1.84}$	&47.13$_{\pm2.49}$	&52.74$_{\pm1.77}$	&59.30$_{\pm1.41}$	&OM	&[\textbf{77.58}$_{\pm1.37}$]	&\textbf{74.91}$_{\pm1.26}$	&\textbf{75.33}$_{\pm2.27}$\\
\hline
\hline
Avg. score	&44.94	&42.74	&43.65	&48.77	&52.19	&44.91	&51.33	&43.78	&-	&\textbf{64.79}	&\textbf{64.07}	&\textbf{64.50}\\
\hline
Avg. rank	&8.80	&9.13	&8.73	&8.00	&6.13	&9.13	&7.00	&8.40	&5.53	&\textbf{2.00}	&\textbf{2.63}	&\textbf{2.37}\\
\hline
\end{tabular}
\begin{tablenotes}
\item[*] Note that OM indicates the out-of-memory error.
\end{tablenotes}
\end{threeparttable}\vskip -0.1in
\end{table*}

\subsection{Comparison with Other Ensemble Clustering Methods }

In this section, we evaluate the proposed ensemble clustering algorithms, namely, MDEC-HC, MDEC-SC, and MDEC-BG, against nine baseline ensemble clustering algorithms. For each test algorithm, the number of clusters is set to the true number of classes on the dataset, which is a commonly-adopted experimental protocol in the literature \cite{liu17_tkde, liu17_bioinformatics}. If an algorithm is not computationally feasible on a dataset due to the out-of-memory error, its score on this dataset will be labeled as "OM". For each  dataset, we run every test algorithm 100 times, and report their average NMI and ARI scores in Tables~\ref{table:compare_ce_nmi} and \ref{table:compare_ce_ari}, respectively.

In terms of NMI, as shown in Table~\ref{table:compare_ce_nmi}, the proposed algorithms outperform the baseline algorithms on most of the benchmark datasets. The SPCE achieves the best NMI scores on 5 out of the 30 datasets, but the three proposed algorithms outperform SPCE on most of the other datasets. It is noteworthy that the proposed MDEC-HC algorithm achieves the best NMI scores on 16 out of the 30 datasets. Also, the proposed MDEC-HC, MDEC-SC, and MDEC-BG algorithms rank in the top three positions 26, 27, and 25 times, respectively, out of the total of 30 datasets, while the fourth best algorithm (i.e., SPCE) ranks in the top three positions just 7 times. In terms of ARI, as shown in Table~\ref{table:compare_ce_nmi}, our three algorithms also exhibit clear advantages over the baselines, ranking in the top three positions 28, 25, and 25 times, respectively, while the fourth best algorithm only ranks in the top three positions 4 times.

Further, to provide a summary view across the 30 benchmark datasets, we report the \emph{average score} and \emph{average rank} of different algorithms in the last two rows in Tables~\ref{table:compare_ce_nmi} and \ref{table:compare_ce_ari}. Note that the average score (across 30 datasets) is computed by taking the average on the NMI (or ARI) scores, while the average rank is obtained by taking the average on the ranking positions, for each algorithm across all the benchmark datasets. Note that if an algorithm is not computationally feasible on all datasets, it will not have an average score, but will still have an average rank, where the \emph{lost} score on a dataset (due to the out-of-memory error) will be considered as the last position on this dataset.
As can be seen in Table~\ref{table:compare_ce_nmi}, the proposed three algorithms achieve  average NMI(\%) scores of $64.79$, $64.07$, and $64.50$, respectively, which are significantly higher than the fourth best average score of $52.19$. In terms of the ranking positions in Table~\ref{table:compare_ce_nmi}, the proposed three algorithms obtain average ranks of $2.00$, $2.63$, and $2.37$, respectively, while the fourth best algorithm only obtains an average rank of $5.53$. Similar advantages can also be observed in terms of ARI. The average ARI scores and the average ranks (across the 30 datasets) of the proposed algorithms significantly outperform the nine baseline algorithms (as shown in Table~\ref{table:compare_ce_ari}).

\begin{table*}[!t]\scriptsize
\centering
\caption{Average performances (w.r.t. ARI) over 100 runs by different ensemble clustering algorithms.
On each dataset, the highest three scores are highlighted in \textbf{bold}, while the highest one in [\textbf{bold and brackets}].}
\label{table:compare_ce_ari}
\begin{tabular}{|m{0.68cm}<{\centering}|m{0.9cm}<{\centering}m{0.97cm}<{\centering}m{0.97cm}<{\centering}m{0.97cm}<{\centering}m{0.97cm}<{\centering}m{0.97cm}<{\centering}m{0.97cm}<{\centering}m{1.16cm}<{\centering}m{1cm}<{\centering}m{1.2cm}<{\centering}m{1.15cm}<{\centering}m{1.2cm}<{\centering}|}
\hline
Dataset &SSCSPA     &SSHGPA     &SSMCLA     &KCC     &PTGP  &SEC    &ECC    &LWGP   &SPCE  &MDEC-HC   &MDEC-SC   &MDEC-BG\\
\hline
\hline
\emph{DS-1}	&8.25$_{\pm1.68}$	&7.01$_{\pm1.67}$	&12.26$_{\pm3.33}$	&13.74$_{\pm7.37}$	&23.96$_{\pm1.57}$	&12.35$_{\pm3.34}$	&22.08$_{\pm1.33}$	&30.83$_{\pm10.37}$	&[\textbf{51.88}$_{\pm4.14}$]	&\textbf{50.42}$_{\pm2.51}$	&\textbf{46.94}$_{\pm2.15}$	&46.71$_{\pm1.81}$\\
\hline
\emph{DS-2}	&36.46$_{\pm2.66}$	&29.64$_{\pm6.81}$	&30.95$_{\pm3.64}$	&35.63$_{\pm2.62}$	&33.18$_{\pm6.35}$	&33.71$_{\pm5.18}$	&35.52$_{\pm4.90}$	&22.98$_{\pm7.25}$	&\textbf{47.03}$_{\pm9.54}$	&[\textbf{53.89}$_{\pm4.35}$]	&46.82$_{\pm2.67}$	&\textbf{46.99}$_{\pm2.42}$\\
\hline
\emph{DS-3}	&65.20$_{\pm3.14}$	&80.99$_{\pm3.44}$	&52.52$_{\pm20.09}$	&69.40$_{\pm4.03}$	&6.57$_{\pm0.00}$	&58.71$_{\pm13.38}$	&71.06$_{\pm1.35}$	&6.57$_{\pm0.00}$	&29.80$_{\pm28.79}$	&[\textbf{92.38}$_{\pm0.00}$]	&[\textbf{92.38}$_{\pm0.00}$]	&[\textbf{92.38}$_{\pm0.00}$]\\
\hline
\emph{DS-4}	&8.97$_{\pm0.50}$	&9.38$_{\pm1.89}$	&\textbf{17.26}$_{\pm6.14}$	&14.44$_{\pm8.72}$	&8.85$_{\pm2.55}$	&11.18$_{\pm7.49}$	&14.64$_{\pm6.83}$	&15.99$_{\pm3.22}$	&[\textbf{21.94}$_{\pm4.88}$]	&\textbf{18.42}$_{\pm0.95}$	&16.89$_{\pm3.03}$	&16.98$_{\pm0.78}$\\
\hline
\emph{DS-5}	&44.73$_{\pm4.34}$	&45.15$_{\pm10.61}$	&73.83$_{\pm8.22}$	&64.01$_{\pm4.32}$	&65.53$_{\pm6.71}$	&57.42$_{\pm5.94}$	&61.72$_{\pm2.50}$	&65.00$_{\pm2.85}$	&59.14$_{\pm0.65}$	&\textbf{80.99}$_{\pm16.82}$	&[\textbf{82.41}$_{\pm9.41}$]	&\textbf{78.53}$_{\pm15.58}$\\
\hline
\emph{DS-6}	&46.12$_{\pm3.25}$	&49.59$_{\pm10.14}$	&39.78$_{\pm18.56}$	&51.89$_{\pm14.45}$	&68.29$_{\pm6.54}$	&60.02$_{\pm9.10}$	&72.63$_{\pm5.64}$	&65.13$_{\pm3.73}$	&56.21$_{\pm5.55}$	&\textbf{79.45}$_{\pm6.05}$	&\textbf{80.14}$_{\pm3.37}$	&[\textbf{80.92}$_{\pm3.17}$]\\
\hline
\emph{DS-7}	&24.54$_{\pm2.60}$	&23.19$_{\pm5.12}$	&20.48$_{\pm5.47}$	&34.18$_{\pm4.19}$	&\textbf{36.43}$_{\pm1.79}$	&35.02$_{\pm3.46}$	&26.20$_{\pm2.62}$	&35.56$_{\pm0.59}$	&32.87$_{\pm0.69}$	&[\textbf{39.86}$_{\pm1.24}$]	&34.12$_{\pm5.93}$	&\textbf{38.57}$_{\pm1.57}$\\
\hline
\emph{DS-8}	&36.14$_{\pm6.23}$	&44.96$_{\pm7.56}$	&24.64$_{\pm13.25}$	&47.44$_{\pm6.69}$	&47.29$_{\pm4.63}$	&42.17$_{\pm10.56}$	&46.25$_{\pm4.74}$	&45.62$_{\pm3.86}$	&50.18$_{\pm3.11}$	&[\textbf{58.47}$_{\pm2.85}$]	&\textbf{57.68}$_{\pm2.86}$	&\textbf{57.04}$_{\pm1.31}$\\
\hline
\emph{DS-9}	&36.17$_{\pm1.94}$	&35.40$_{\pm4.20}$	&28.27$_{\pm11.94}$	&42.58$_{\pm3.68}$	&41.44$_{\pm4.38}$	&42.03$_{\pm5.91}$	&40.05$_{\pm4.08}$	&38.63$_{\pm2.83}$	&38.14$_{\pm8.43}$	&\textbf{52.65}$_{\pm2.57}$	&\textbf{55.98}$_{\pm1.60}$	&[\textbf{56.48}$_{\pm2.35}$]\\
\hline
\emph{DS-10}	&35.96$_{\pm5.65}$	&[\textbf{43.82}$_{\pm1.40}$]	&30.87$_{\pm5.69}$	&38.30$_{\pm3.28}$	&38.67$_{\pm1.66}$	&33.22$_{\pm6.87}$	&38.75$_{\pm0.00}$	&0.01$_{\pm0.00}$	&25.41$_{\pm10.12}$	&39.53$_{\pm4.62}$	&\textbf{40.12}$_{\pm2.49}$	&\textbf{42.19}$_{\pm2.65}$\\
\hline
\emph{DS-11}	&12.36$_{\pm0.14}$	&8.65$_{\pm3.13}$	&13.12$_{\pm8.07}$	&88.94$_{\pm14.60}$	&93.09$_{\pm1.10}$	&56.16$_{\pm49.17}$	&93.54$_{\pm3.52}$	&0.22$_{\pm12.44}$	&53.28$_{\pm35.18}$	&[\textbf{97.90}$_{\pm1.04}$]	&\textbf{97.27}$_{\pm1.13}$	&\textbf{97.60}$_{\pm0.94}$\\
\hline
\emph{DS-12}	&23.12$_{\pm2.21}$	&24.41$_{\pm4.01}$	&6.17$_{\pm5.31}$	&18.46$_{\pm2.35}$	&18.15$_{\pm5.31}$	&15.03$_{\pm3.78}$	&20.46$_{\pm3.96}$	&17.53$_{\pm0.64}$	&17.68$_{\pm1.57}$	&\textbf{30.83}$_{\pm6.96}$	&\textbf{32.85}$_{\pm5.88}$	&[\textbf{41.49}$_{\pm3.12}$]\\
\hline
\emph{DS-13}	&43.00$_{\pm0.74}$	&39.27$_{\pm2.45}$	&43.39$_{\pm2.01}$	&55.44$_{\pm23.52}$	&\textbf{97.68}$_{\pm9.43}$	&48.03$_{\pm17.25}$	&61.88$_{\pm26.89}$	&76.94$_{\pm9.01}$	&94.16$_{\pm6.26}$	&[\textbf{99.00}$_{\pm2.08}$]	&\textbf{96.88}$_{\pm2.61}$	&95.98$_{\pm2.27}$\\
\hline
\emph{DS-14}	&26.77$_{\pm4.98}$	&33.34$_{\pm3.38}$	&35.32$_{\pm3.29}$	&40.71$_{\pm7.03}$	&41.90$_{\pm2.85}$	&31.78$_{\pm7.48}$	&41.49$_{\pm1.70}$	&42.84$_{\pm3.49}$	&[\textbf{50.15}$_{\pm1.88}$]	&\textbf{45.52}$_{\pm1.30}$	&\textbf{43.09}$_{\pm1.23}$	&42.87$_{\pm1.22}$\\
\hline
\emph{DS-15}	&68.50$_{\pm2.50}$	&48.65$_{\pm7.02}$	&65.12$_{\pm3.81}$	&46.04$_{\pm6.69}$	&76.81$_{\pm6.68}$	&34.81$_{\pm8.69}$	&72.54$_{\pm7.32}$	&56.48$_{\pm10.58}$	&65.09$_{\pm12.86}$	&[\textbf{84.57}$_{\pm2.99}$]	&\textbf{84.10}$_{\pm2.24}$	&\textbf{82.13}$_{\pm2.33}$\\
\hline
\emph{DS-16}	&28.99$_{\pm1.67}$	&36.03$_{\pm1.93}$	&7.45$_{\pm3.67}$	&17.68$_{\pm4.70}$	&6.17$_{\pm1.50}$	&10.47$_{\pm4.51}$	&16.87$_{\pm7.24}$	&3.07$_{\pm0.38}$	&6.86$_{\pm0.68}$	&[\textbf{57.54}$_{\pm3.33}$]	&\textbf{48.82}$_{\pm2.27}$	&\textbf{48.74}$_{\pm3.14}$\\
\hline
\emph{DS-17}	&17.65$_{\pm6.12}$	&\textbf{21.11}$_{\pm3.90}$	&5.48$_{\pm9.14}$	&10.69$_{\pm5.22}$	&13.90$_{\pm3.04}$	&12.29$_{\pm5.34}$	&14.27$_{\pm3.02}$	&-6.19$_{\pm0.48}$	&8.74$_{\pm2.76}$	&[\textbf{22.16}$_{\pm3.96}$]	&18.87$_{\pm2.57}$	&\textbf{19.48}$_{\pm2.14}$\\
\hline
\emph{DS-18}	&24.15$_{\pm1.51}$	&24.24$_{\pm3.64}$	&29.01$_{\pm4.07}$	&30.23$_{\pm5.14}$	&29.89$_{\pm3.72}$	&22.30$_{\pm4.04}$	&25.08$_{\pm2.68}$	&27.22$_{\pm2.30}$	&26.13$_{\pm2.58}$	&[\textbf{51.33}$_{\pm4.08}$]	&\textbf{40.39}$_{\pm3.17}$	&\textbf{41.91}$_{\pm2.49}$\\
\hline
\emph{DS-19}	&46.24$_{\pm2.84}$	&33.48$_{\pm2.31}$	&\textbf{53.59}$_{\pm1.74}$	&44.11$_{\pm2.07}$	&46.03$_{\pm2.12}$	&43.64$_{\pm2.18}$	&44.63$_{\pm1.85}$	&\textbf{54.18}$_{\pm1.68}$	&44.69$_{\pm5.44}$	&[\textbf{56.57}$_{\pm1.90}$]	&49.05$_{\pm0.60}$	&50.49$_{\pm1.68}$\\
\hline
\emph{DS-20}	&64.16$_{\pm1.84}$	&59.10$_{\pm3.94}$	&56.54$_{\pm9.01}$	&52.50$_{\pm3.59}$	&37.76$_{\pm4.80}$	&51.94$_{\pm5.44}$	&54.40$_{\pm2.30}$	&64.33$_{\pm1.99}$	&64.99$_{\pm3.39}$	&[\textbf{81.58}$_{\pm1.30}$]	&\textbf{77.21}$_{\pm2.72}$	&\textbf{80.21}$_{\pm1.68}$\\
\hline
\emph{DS-21}	&39.90$_{\pm1.88}$	&41.42$_{\pm2.85}$	&43.39$_{\pm2.54}$	&38.72$_{\pm1.92}$	&37.71$_{\pm2.89}$	&35.75$_{\pm5.65}$	&39.99$_{\pm3.26}$	&42.29$_{\pm2.83}$	&42.64$_{\pm4.96}$	&\textbf{44.16}$_{\pm1.61}$	&[\textbf{46.13}$_{\pm2.49}$]	&\textbf{45.08}$_{\pm1.85}$\\
\hline
\emph{DS-22}	&25.06$_{\pm1.23}$	&15.59$_{\pm1.15}$	&24.82$_{\pm1.88}$	&25.01$_{\pm1.21}$	&27.02$_{\pm0.52}$	&27.16$_{\pm0.77}$	&25.46$_{\pm1.04}$	&24.94$_{\pm0.94}$	&24.21$_{\pm2.45}$	&\textbf{30.56}$_{\pm1.61}$	&\textbf{34.27}$_{\pm0.90}$	&[\textbf{34.62}$_{\pm0.93}$]\\
\hline
\emph{DS-23}	&0.63$_{\pm0.09}$	&1.05$_{\pm0.24}$	&0.91$_{\pm0.14}$	&0.61$_{\pm0.07}$	&0.84$_{\pm0.09}$	&0.80$_{\pm0.12}$	&0.64$_{\pm0.15}$	&0.72$_{\pm0.06}$	&0.41$_{\pm0.08}$	&\textbf{7.77}$_{\pm0.58}$	&\textbf{11.23}$_{\pm0.54}$	&[\textbf{13.11}$_{\pm0.89}$]\\
\hline
\emph{DS-24}	&45.44$_{\pm4.83}$	&21.30$_{\pm1.95}$	&46.18$_{\pm3.41}$	&39.94$_{\pm3.42}$	&48.25$_{\pm1.13}$	&39.33$_{\pm4.61}$	&38.02$_{\pm3.31}$	&51.21$_{\pm3.99}$	&49.28$_{\pm3.89}$	&[\textbf{60.23}$_{\pm1.13}$]	&\textbf{53.71}$_{\pm2.35}$	&\textbf{55.74}$_{\pm2.64}$\\
\hline
\emph{DS-25}	&27.98$_{\pm1.33}$	&35.84$_{\pm3.76}$	&27.62$_{\pm4.08}$	&30.77$_{\pm2.06}$	&31.17$_{\pm1.11}$	&29.95$_{\pm1.88}$	&31.24$_{\pm1.45}$	&33.48$_{\pm1.69}$	&37.60$_{\pm3.33}$	&[\textbf{67.41}$_{\pm2.82}$]	&\textbf{61.32}$_{\pm6.10}$	&\textbf{63.91}$_{\pm5.02}$\\
\hline
\emph{DS-26}	&78.94$_{\pm4.63}$	&61.55$_{\pm4.35}$	&84.82$_{\pm3.40}$	&62.73$_{\pm6.98}$	&79.62$_{\pm4.36}$	&48.52$_{\pm8.63}$	&65.22$_{\pm2.48}$	&85.68$_{\pm2.94}$	&77.17$_{\pm7.15}$	&\textbf{88.10}$_{\pm4.07}$	&[\textbf{94.46}$_{\pm0.44}$]	&\textbf{94.19}$_{\pm0.40}$\\
\hline
\emph{DS-27}	&7.98$_{\pm0.40}$	&6.88$_{\pm0.85}$	&8.97$_{\pm0.51}$	&9.51$_{\pm0.32}$	&9.34$_{\pm0.24}$	&9.22$_{\pm0.37}$	&\textbf{9.70}$_{\pm0.44}$	&9.69$_{\pm0.47}$	&4.03$_{\pm0.59}$	&9.48$_{\pm0.38}$	&\textbf{10.08}$_{\pm0.30}$	&[\textbf{10.17}$_{\pm0.35}$]\\
\hline
\emph{DS-28}	&42.13$_{\pm4.44}$	&25.53$_{\pm3.13}$	&46.71$_{\pm3.13}$	&40.52$_{\pm3.60}$	&53.84$_{\pm2.85}$	&40.97$_{\pm3.11}$	&40.84$_{\pm3.37}$	&48.54$_{\pm1.94}$	&34.18$_{\pm12.43}$	&[\textbf{70.13}$_{\pm4.54}$]	&\textbf{65.83}$_{\pm2.05}$	&\textbf{69.39}$_{\pm1.88}$\\
\hline
\emph{DS-29}	&26.82$_{\pm8.18}$	&24.92$_{\pm13.52}$	&58.48$_{\pm5.61}$	&50.80$_{\pm11.04}$	&60.54$_{\pm0.92}$	&11.89$_{\pm6.18}$	&56.48$_{\pm2.15}$	&6.77$_{\pm17.60}$	&5.27$_{\pm2.76}$	&\textbf{76.30}$_{\pm4.91}$	&\textbf{77.15}$_{\pm0.64}$	&[\textbf{77.59}$_{\pm0.29}$]\\
\hline
\emph{DS-30}	&33.95$_{\pm4.16}$	&19.23$_{\pm1.81}$	&38.28$_{\pm3.47}$	&33.41$_{\pm1.81}$	&52.57$_{\pm3.11}$	&30.04$_{\pm3.59}$	&37.33$_{\pm2.13}$	&43.99$_{\pm1.67}$	&OM	&[\textbf{67.35}$_{\pm2.13}$]	&\textbf{61.86}$_{\pm2.41}$	&\textbf{62.52}$_{\pm3.90}$\\
\hline
\hline
Avg. score	&34.21	&31.69	&34.21	&38.28	&41.08	&32.86	&40.63	&33.68	&-	&\textbf{57.15}	&\textbf{55.27}	&\textbf{56.13}\\
\hline
Avg. rank	&8.60	&8.87	&8.23	&7.77	&6.57	&9.03	&7.00	&7.33	&7.53	&\textbf{1.93}	&\textbf{2.73}	&\textbf{2.27}\\
\hline
\end{tabular}\vskip -0.1in
\end{table*}

\subsection{Robustness to Ensemble Sizes}
\label{sec:compEnSize}

\begin{figure*}[!th]
\begin{center}
{\subfigure
{\includegraphics[width=0.291\columnwidth]{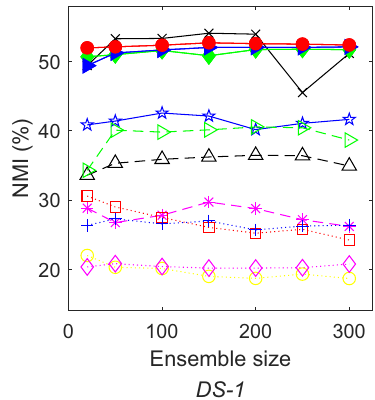}}}
{\subfigure
{\includegraphics[width=0.291\columnwidth]{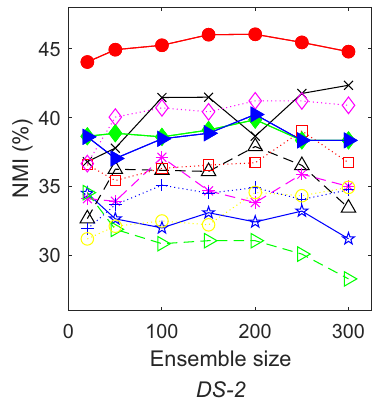}}}
{\subfigure
{\includegraphics[width=0.291\columnwidth]{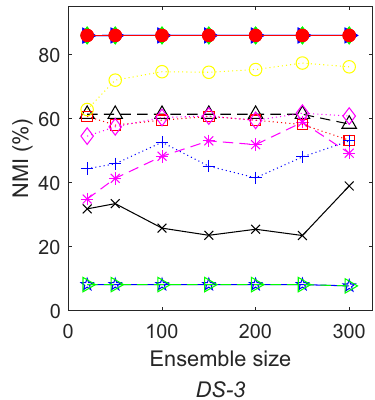}}}
{\subfigure
{\includegraphics[width=0.291\columnwidth]{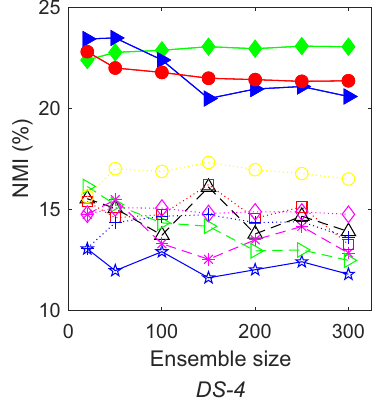}}}
{\subfigure
{\includegraphics[width=0.291\columnwidth]{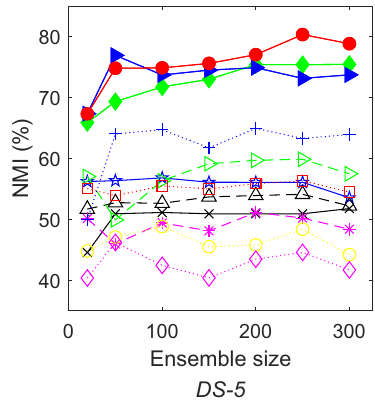}}}
{\subfigure
{\includegraphics[width=0.291\columnwidth]{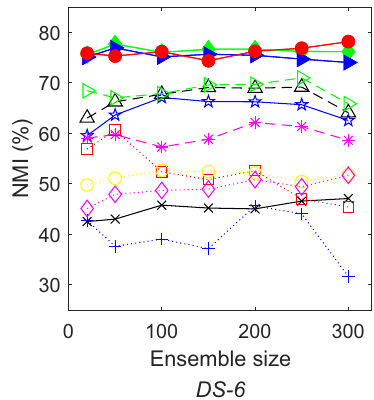}}}
{\subfigure
{\includegraphics[width=0.291\columnwidth]{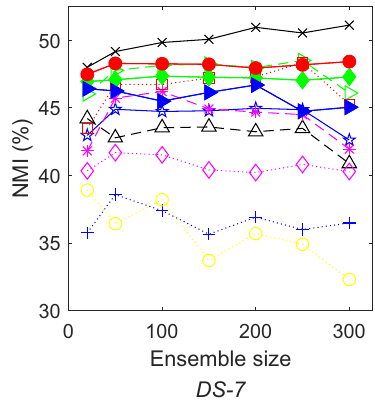}}}
{\subfigure
{\includegraphics[width=0.291\columnwidth]{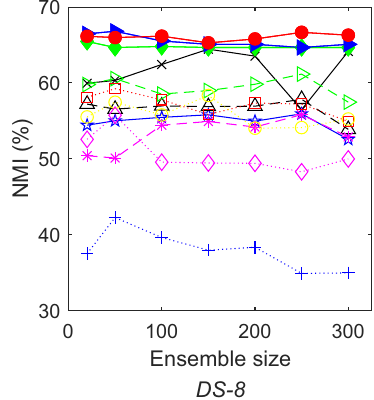}}}
{\subfigure
{\includegraphics[width=0.291\columnwidth]{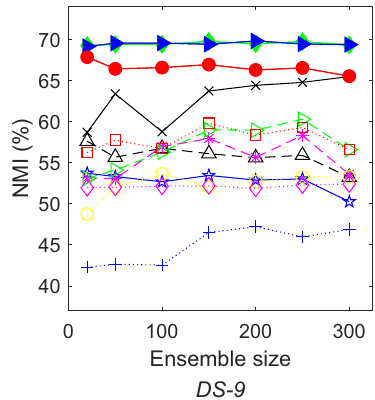}}}
{\subfigure
{\includegraphics[width=0.291\columnwidth]{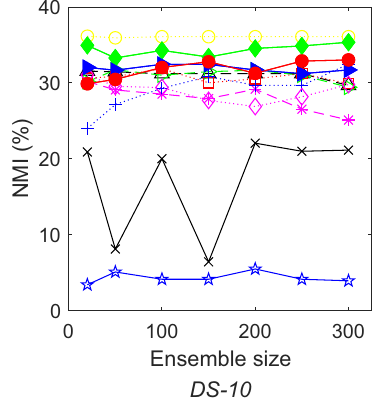}}}
{\subfigure
{\includegraphics[width=0.291\columnwidth]{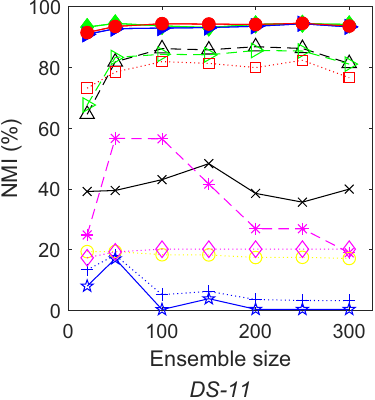}}}
{\subfigure
{\includegraphics[width=0.291\columnwidth]{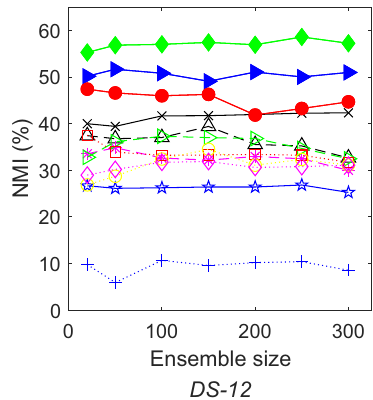}}}
{\subfigure
{\includegraphics[width=0.291\columnwidth]{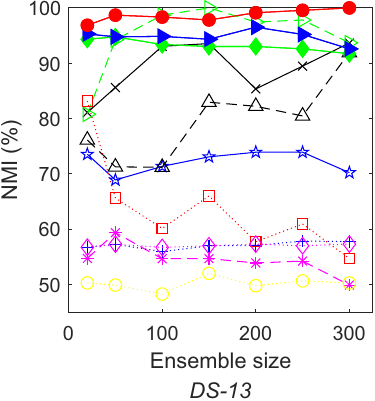}}}
{\subfigure
{\includegraphics[width=0.291\columnwidth]{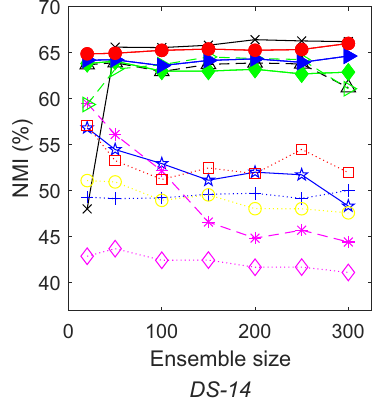}}}
{\subfigure
{\includegraphics[width=0.291\columnwidth]{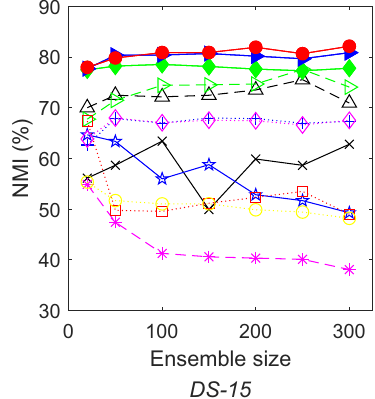}}}
{\subfigure
{\includegraphics[width=0.291\columnwidth]{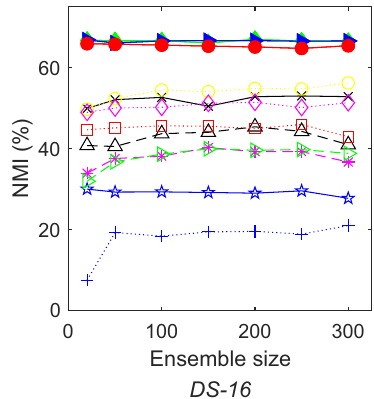}}}
{\subfigure
{\includegraphics[width=0.291\columnwidth]{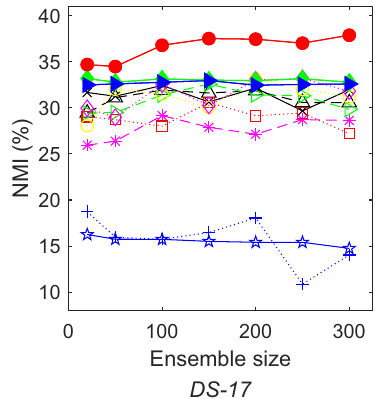}}}
{\subfigure
{\includegraphics[width=0.291\columnwidth]{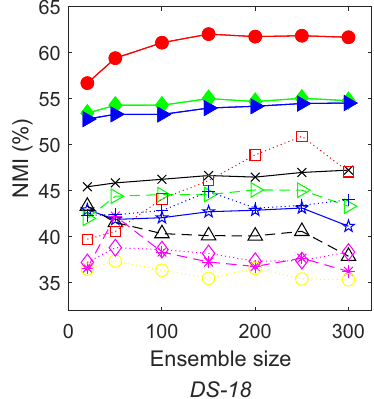}}}
{\subfigure
{\includegraphics[width=0.291\columnwidth]{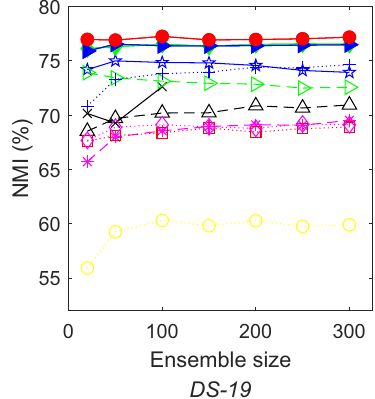}}}
{\subfigure
{\includegraphics[width=0.291\columnwidth]{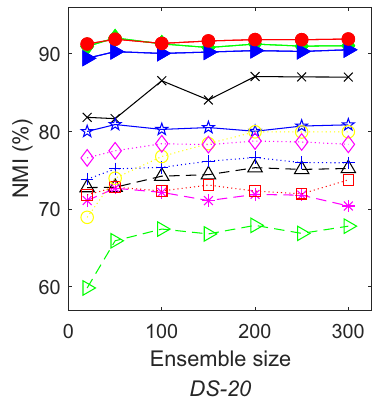}}}
{\subfigure
{\includegraphics[width=0.291\columnwidth]{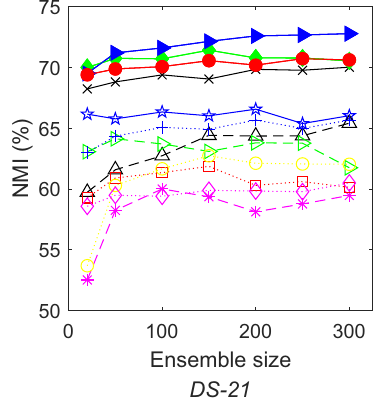}}}
{\subfigure
{\includegraphics[width=0.291\columnwidth]{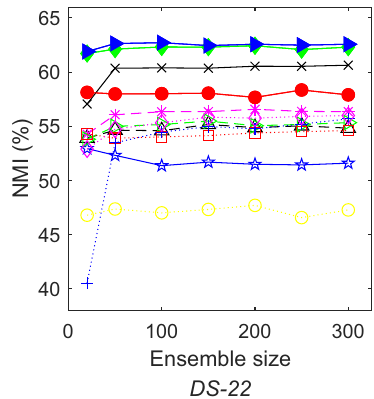}}}
{\subfigure
{\includegraphics[width=0.291\columnwidth]{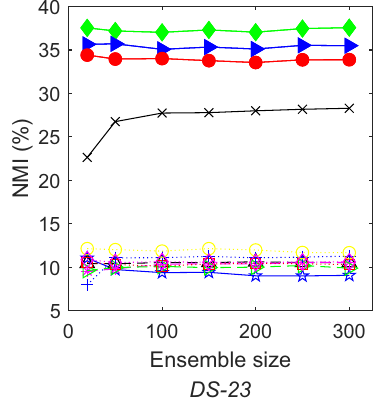}}}
{\subfigure
{\includegraphics[width=0.291\columnwidth]{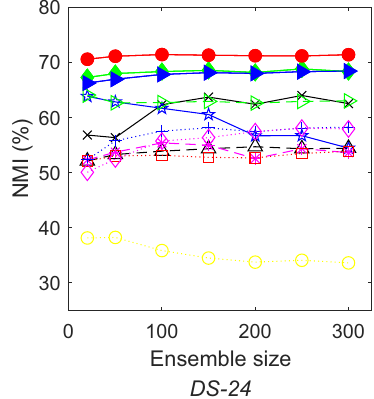}}}
{\subfigure
{\includegraphics[width=0.291\columnwidth]{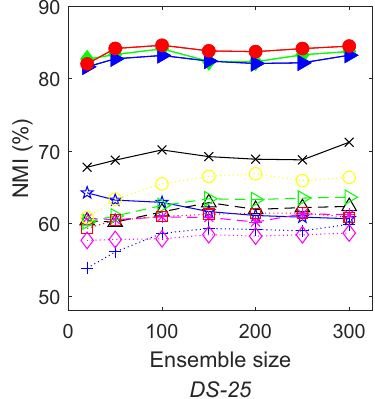}}}
{\subfigure
{\includegraphics[width=0.291\columnwidth]{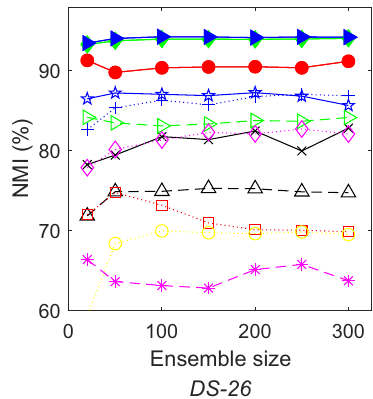}}}
{\subfigure
{\includegraphics[width=0.291\columnwidth]{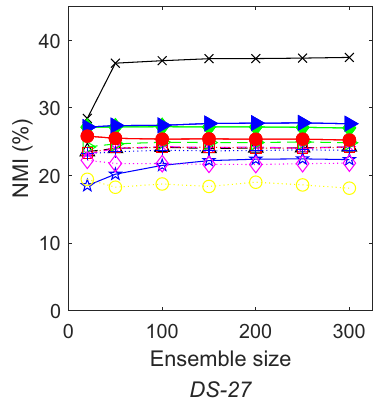}}}
{\subfigure
{\includegraphics[width=0.291\columnwidth]{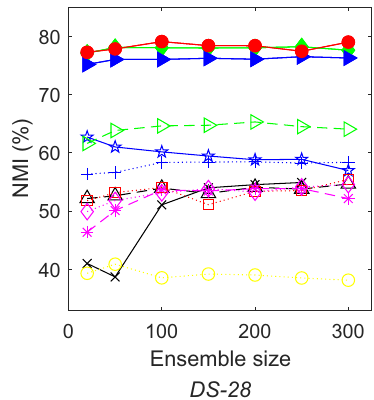}}}
{\subfigure
{\includegraphics[width=0.291\columnwidth]{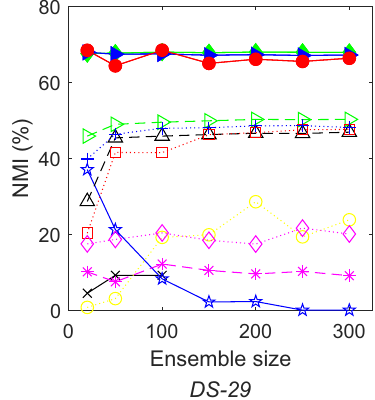}}}
{\subfigure
{\includegraphics[width=0.291\columnwidth]{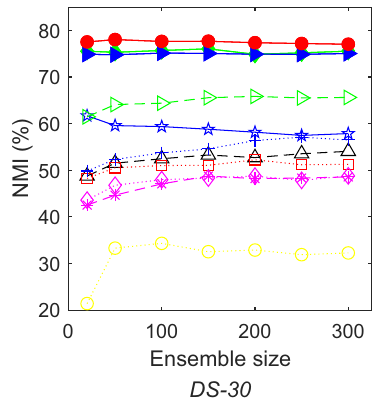}}}
{\subfigure
{\includegraphics[width=1.85\columnwidth]{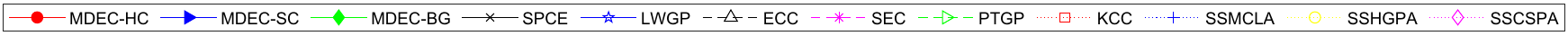}}}
\caption{The average performances (w.r.t. NMI) over 20 runs by different ensemble clustering algorithms with varying ensemble sizes $M$. Note that if an algorithm is not feasible (due to the out-of-memory error) for some large ensemble sizes on a dataset, then its curve will stop earlier than other algorithms.}
\label{fig:comp_nmi_Msize}
\end{center}
\end{figure*}

\begin{figure*}[!th]
\begin{center}
{\subfigure
{\includegraphics[width=0.291\columnwidth]{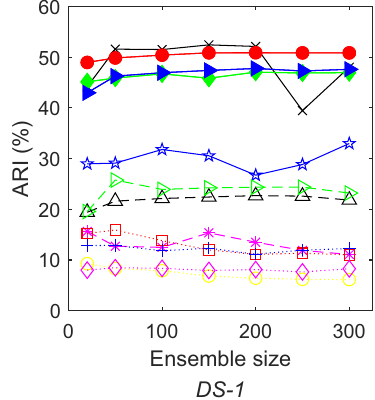}}}
{\subfigure
{\includegraphics[width=0.291\columnwidth]{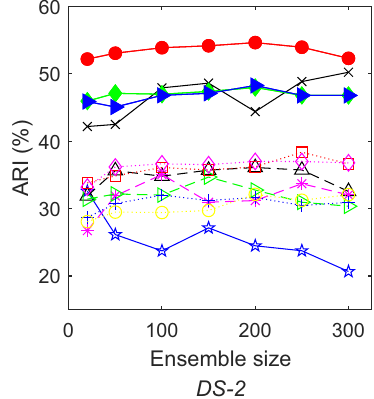}}}
{\subfigure
{\includegraphics[width=0.291\columnwidth]{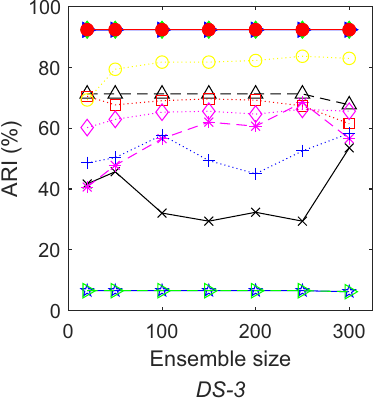}}}
{\subfigure
{\includegraphics[width=0.291\columnwidth]{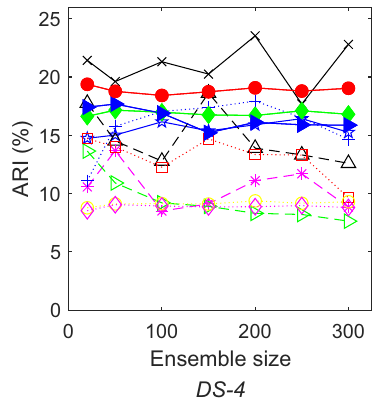}}}
{\subfigure
{\includegraphics[width=0.291\columnwidth]{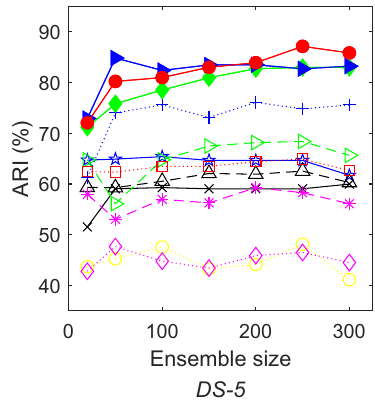}}}
{\subfigure
{\includegraphics[width=0.291\columnwidth]{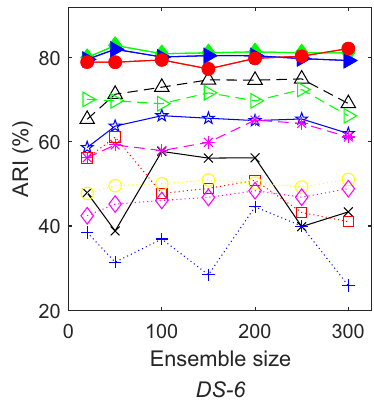}}}
{\subfigure
{\includegraphics[width=0.291\columnwidth]{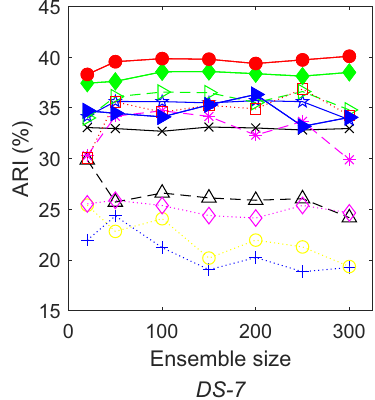}}}
{\subfigure
{\includegraphics[width=0.291\columnwidth]{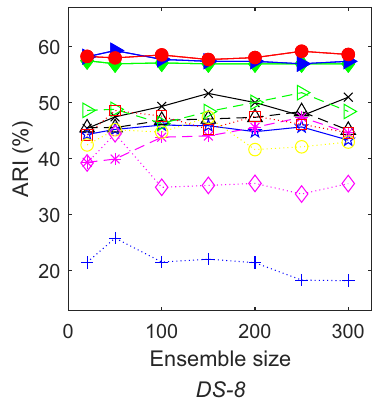}}}
{\subfigure
{\includegraphics[width=0.291\columnwidth]{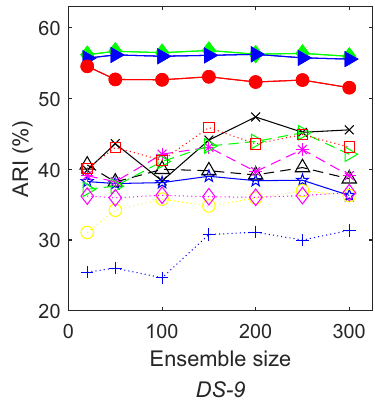}}}
{\subfigure
{\includegraphics[width=0.291\columnwidth]{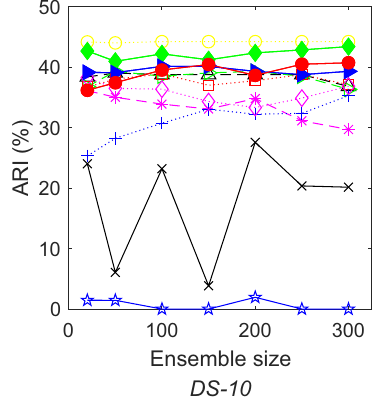}}}
{\subfigure
{\includegraphics[width=0.291\columnwidth]{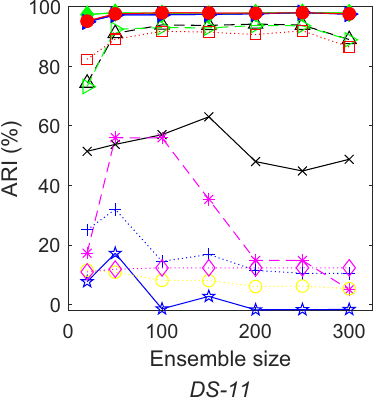}}}
{\subfigure
{\includegraphics[width=0.291\columnwidth]{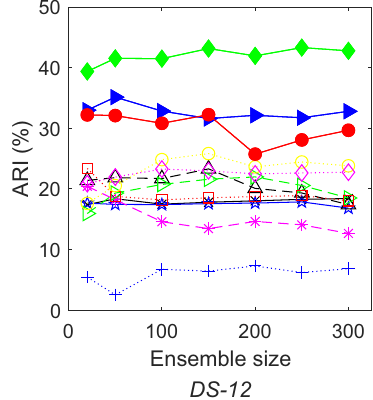}}}
{\subfigure
{\includegraphics[width=0.291\columnwidth]{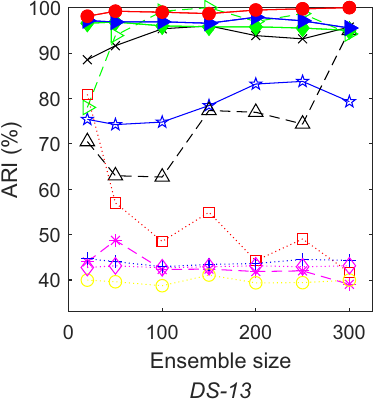}}}
{\subfigure
{\includegraphics[width=0.291\columnwidth]{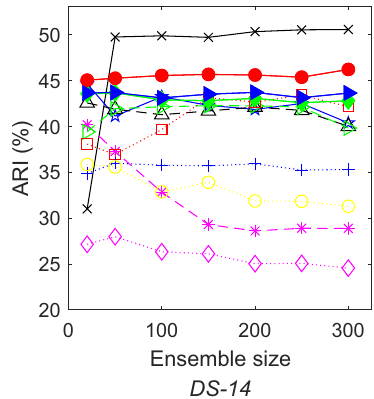}}}
{\subfigure
{\includegraphics[width=0.291\columnwidth]{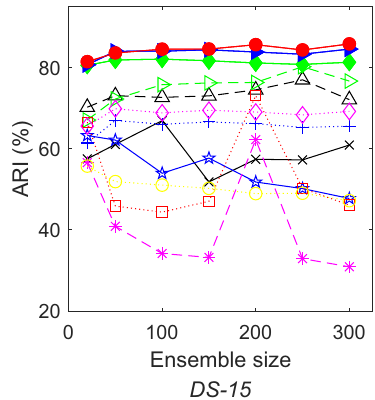}}}
{\subfigure
{\includegraphics[width=0.291\columnwidth]{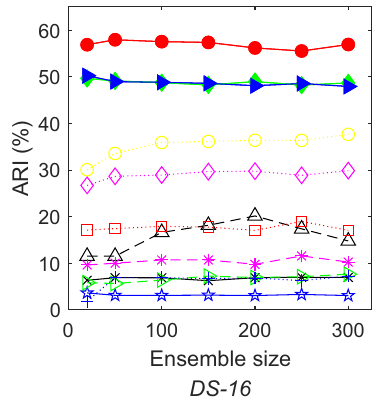}}}
{\subfigure
{\includegraphics[width=0.291\columnwidth]{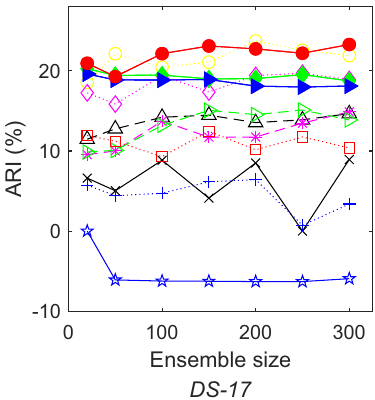}}}
{\subfigure
{\includegraphics[width=0.291\columnwidth]{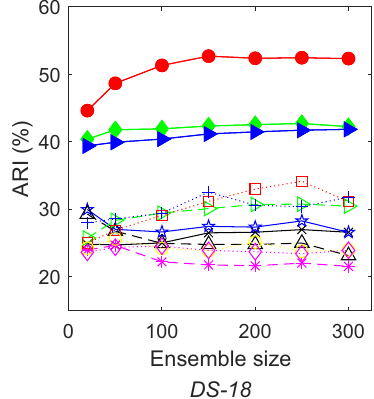}}}
{\subfigure
{\includegraphics[width=0.291\columnwidth]{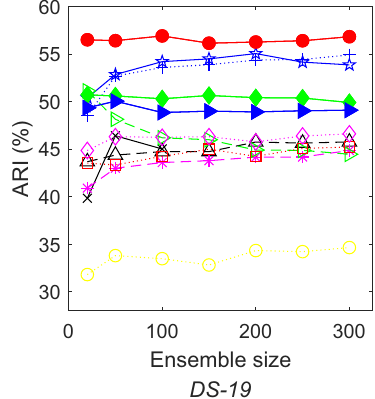}}}
{\subfigure
{\includegraphics[width=0.291\columnwidth]{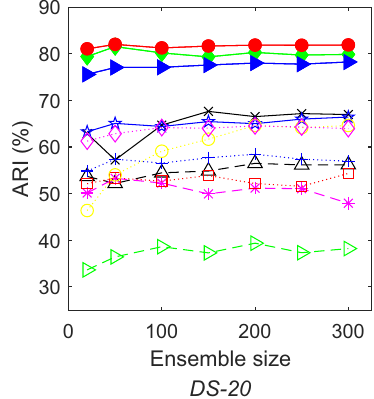}}}
{\subfigure
{\includegraphics[width=0.291\columnwidth]{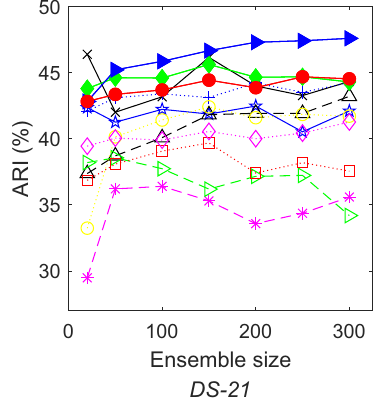}}}
{\subfigure
{\includegraphics[width=0.291\columnwidth]{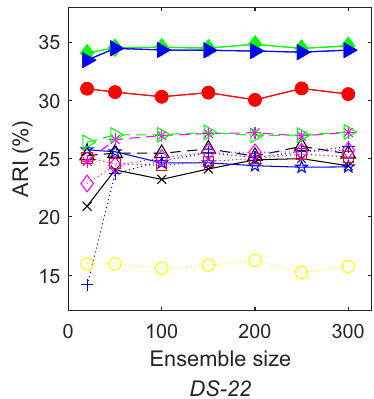}}}
{\subfigure
{\includegraphics[width=0.291\columnwidth]{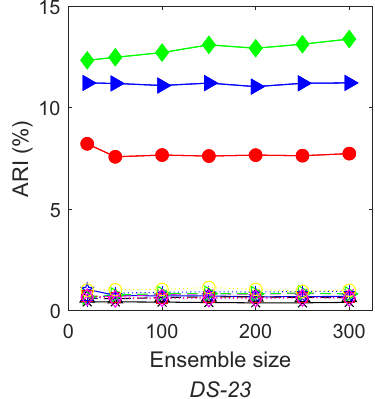}}}
{\subfigure
{\includegraphics[width=0.291\columnwidth]{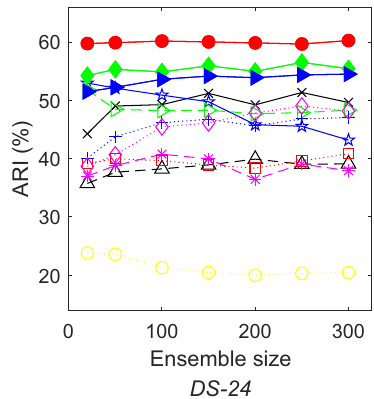}}}
{\subfigure
{\includegraphics[width=0.291\columnwidth]{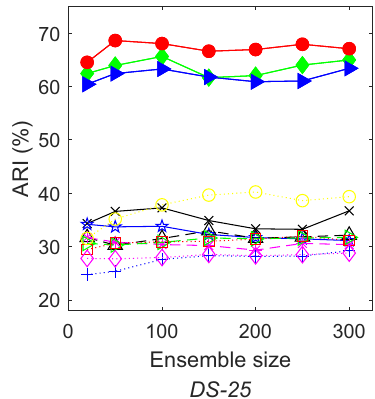}}}
{\subfigure
{\includegraphics[width=0.291\columnwidth]{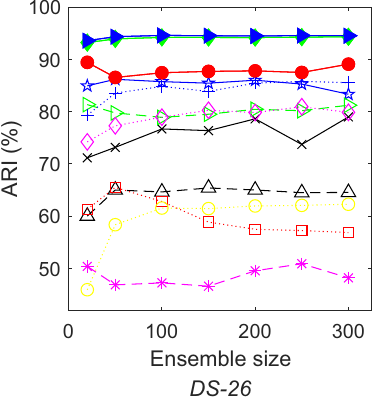}}}
{\subfigure
{\includegraphics[width=0.291\columnwidth]{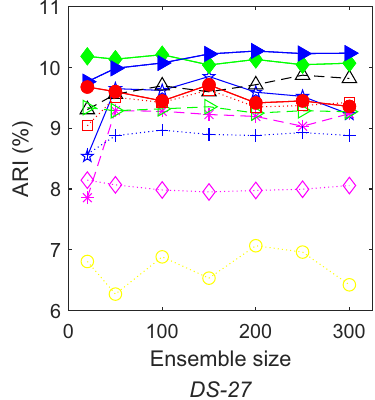}}}
{\subfigure
{\includegraphics[width=0.291\columnwidth]{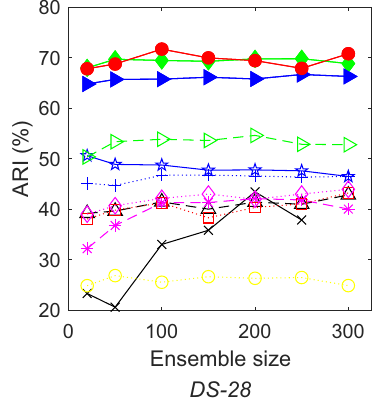}}}
{\subfigure
{\includegraphics[width=0.291\columnwidth]{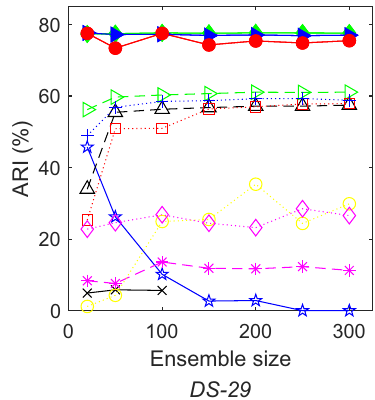}}}
{\subfigure
{\includegraphics[width=0.291\columnwidth]{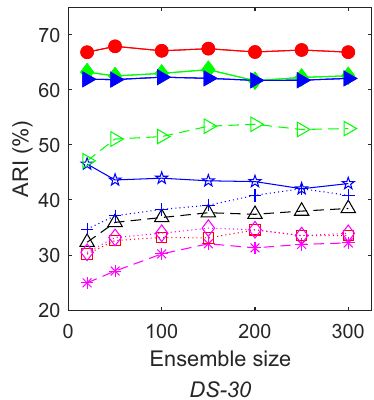}}}
{\subfigure
{\includegraphics[width=1.85\columnwidth]{Figures/compVaryMsizeII_legend}}}
\caption{The average performances (w.r.t. ARI) over 20 runs by different ensemble clustering algorithms with varying ensemble sizes $M$. Note that if an algorithm is not feasible (due to the out-of-memory error) for some large ensemble sizes on a dataset, then its curve will stop earlier than other algorithms.}
\label{fig:comp_ari_Msize}
\end{center}
\end{figure*}

In this section, we compare the performances of different ensemble clustering algorithms with varying ensemble sizes. Specifically, we perform the test algorithms on the datasets with the ensemble size varying from $20$ to $300$, and report their average NMI and ARI scores in Figs.~\ref{fig:comp_nmi_Msize} and \ref{fig:comp_ari_Msize}, respectively.

In terms of NMI, as can be seen in Fig.~\ref{fig:comp_nmi_Msize}, the proposed algorithms yield stably high
performances across the 30 benchmark datasets, with varying ensemble sizes. Although the SPCE algorithm outperforms our algorithms on the \emph{DS-7} and \emph{DS-27} datasets, yet on most of the other datasets our algorithms achieve better or comparable performance when compared to the baseline ensemble clustering algorithms. Especially, on the \emph{DS-3}, \emph{DS-4}, \emph{DS-5}, \emph{DS-6}, \emph{DS-9}, \emph{DS-16}, \emph{DS-18}, \emph{DS-20}, \emph{DS-23}, \emph{DS-24}, \emph{DS-25}, \emph{DS-26}, \emph{DS-28}, \emph{DS-29}, and \emph{DS-30} datasets, our three proposed algorithms have shown clear advantages over the baseline algorithms as the ensemble size goes from 20 to 300. Similarly, in terms of ARI, our algorithms also exhibit very competitive performances on most of the benchmark datasets with varying ensemble sizes (as can be seen in Fig.~\ref{fig:comp_ari_Msize}).

\begin{figure}[!t]\vskip -0.1in
\begin{center}
{\subfigure[NMI]
{\includegraphics[width=0.35\linewidth]{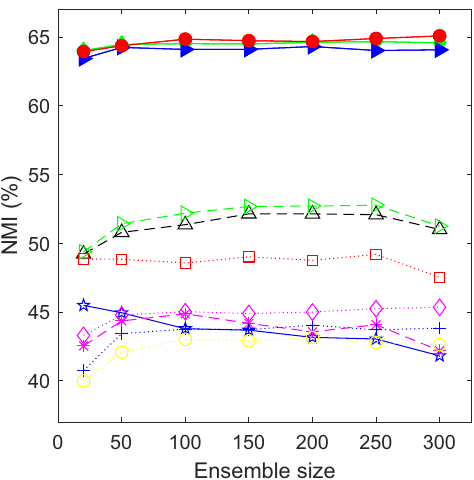}\label{fig:compVaryM_nmi}}}
{\subfigure[ARI]
{\includegraphics[width=0.35\linewidth]{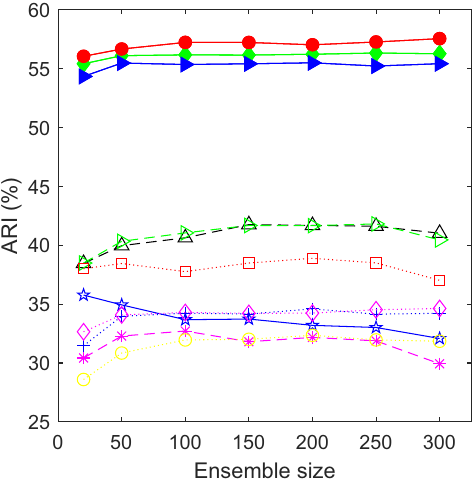}\label{fig:compVaryM_ari}}}
{
{\includegraphics[width=0.175\linewidth]{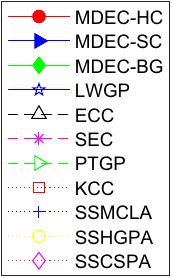}}}
\caption{Average curves (across 30 datasets) by different algorithms with varying ensemble size $M$. Note that Fig.~\ref{fig:compVaryM_nmi} is obtained by averaging the 30 sub-figures in Figs.~\ref{fig:comp_nmi_Msize}, and Fig.~\ref{fig:compVaryM_ari} the 30 sub-figures in Figs.~\ref{fig:comp_ari_Msize}.}\vskip -0.15in
\label{fig:compVaryM_all} 
\end{center}
\end{figure}

Further, in Fig.~\ref{fig:compVaryM_all}, we illustrates the average NMI and ARI curves (across the 30 benchmark datasets) by different ensemble clustering algorithms with varying ensemble sizes. Specifically, Fig.~\ref{fig:compVaryM_nmi} is obtained by taking the average of the 30 sub-figures in Fig.~\ref{fig:comp_nmi_Msize}, while Fig.~\ref{fig:compVaryM_ari} by taking the average of the 30 sub-figures in Fig.~\ref{fig:comp_ari_Msize}. As can be observed in Fig.~\ref{fig:compVaryM_all}, the proposed algorithms achieve significantly better performances (w.r.t. both NMI and ARI) than the baseline ensemble clustering algorithms across the 30 benchmark datasets. Even when compared to the best two baseline algorithms, i.e., PTGP and ECC, the proposed three algorithms can still consistently achieve approximately 20\% higher average NMI/ARI scores.

\subsection{Influence of Metrics, Subspaces, and Clusters}
\label{sec:cmpComponent}

In this paper, we jointly exploit large populations of diversified metrics, random subspaces, and weighted clusters in a unified ensemble clustering framework, and propose three specific ensemble clustering algorithms, termed MDEC-HC, MDEC-SC, and MDEC-BG, respectively. In this section, we will evaluate the influence of the three key factors (i.e., diversified metrics, random subspaces, and weighted clusters) upon the proposed three algorithms.

\begin{figure}[!t]
\begin{center}
{\subfigure[MDEC-HC]
{\includegraphics[width=0.21\linewidth]{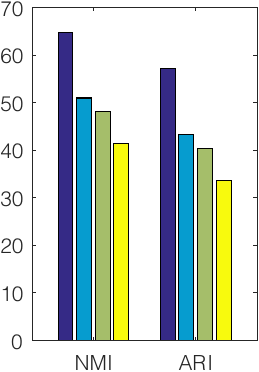}}}
{\subfigure[MDEC-SC]
{\includegraphics[width=0.21\linewidth]{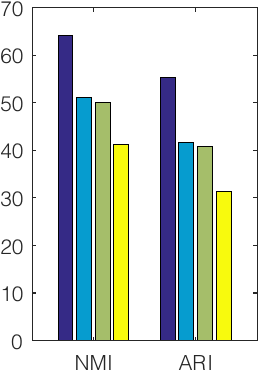}}}
{\subfigure[MDEC-BG]
{\includegraphics[width=0.21\linewidth]{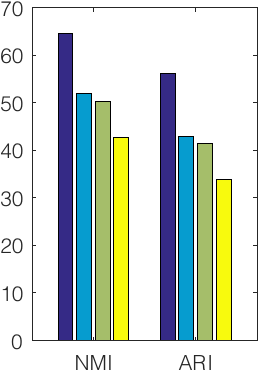}}}
{\subfigure
{\includegraphics[width=0.23\linewidth]{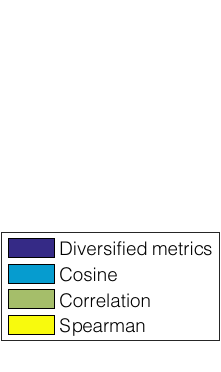}}}
\caption{Average performance (across 30 datasets) of the proposed algorithms using diversified metrics against using other similarity metrics.}
\label{fig:compMetric}
\end{center}
\end{figure}

First, we compare the diversified metrics with several widely used similarity metrics, namely, cosine similarity, correlation coefficient, and Spearman correlation coefficient. For each of the three proposed algorithms, we test its performance of using the diversified metrics against the above three metrics. Taking MDEC-HC as an example, as shown in Fig.~\ref{fig:compMetric}, using diversified metrics leads to an average NMI(\%) score $67.49$, while using the three conventional similarity metrics leads to average NMI(\%) scores of $50.99$, $48.23$, and $41.34$, respectively. Similar advantages of using diversified metrics can also be observed in the cases of MDEC-SC and MDEC-BG, which confirm that the use of diversified metrics in the proposed algorithms can substantially benefits the consensus clustering performance (as shown in Fig.~\ref{fig:compMetric}).

\begin{figure}[!t]
\begin{center}
{\subfigure[MDEC-HC]
{\includegraphics[width=0.278\linewidth]{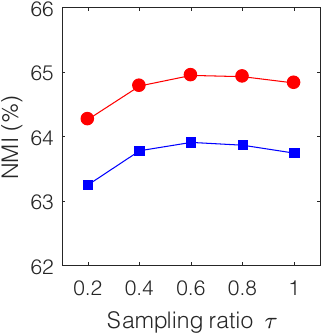}\label{fig:compWCls_VaryM_nmi_HC}}}
{\subfigure[MDEC-SC]
{\includegraphics[width=0.278\linewidth]{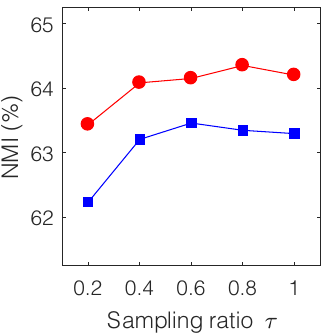}\label{fig:compWCls_VaryM_nmi_SC}}}
{\subfigure[MDEC-BG]
{\includegraphics[width=0.278\linewidth]{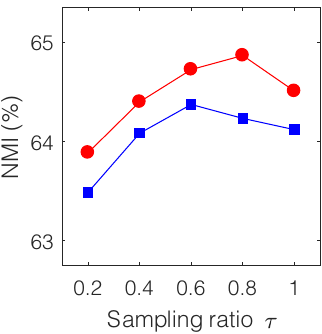}\label{fig:compWCls_VaryM_nmi_BG}}}
{\subfigure
{\includegraphics[width=0.67\columnwidth]{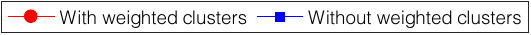}}}
\caption{Average NMI scores (across 30 datasets) of the proposed algorithms \emph{with} and \emph{without} weighted clusters using varying sampling ratios $\tau$.}
\label{fig:compRatioLW_nmi}
\end{center}
\end{figure}

\begin{figure}[!t]
\begin{center}
{\subfigure[MDEC-HC]
{\includegraphics[width=0.278\linewidth]{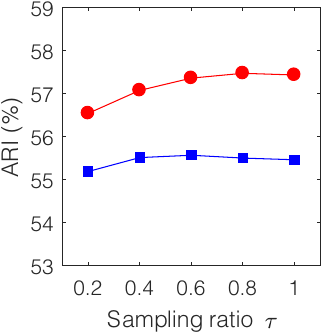}\label{fig:compWCls_VaryM_ari_HC}}}
{\subfigure[MDEC-SC]
{\includegraphics[width=0.278\linewidth]{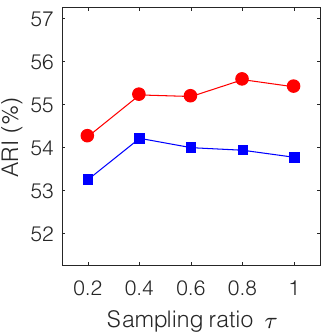}\label{fig:compWCls_VaryM_ari_SC}}}
{\subfigure[MDEC-BG]
{\includegraphics[width=0.278\linewidth]{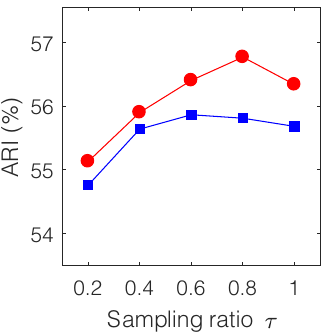}\label{fig:compWCls_VaryM_ari_BG}}}
{\subfigure
{\includegraphics[width=0.67\columnwidth]{Figures/compVaryRatio_legend}}}
\caption{Average ARI scores (across 30 datasets) of the proposed algorithms \emph{with} and \emph{without} weighted clusters using varying sampling ratios $\tau$.}
\label{fig:compRatioLW_ari}
\end{center}
\end{figure}

Second, we evaluate the performances of the proposed ensemble clustering algorithms with different subspace sampling ratio $\tau$, which varies from $0.1$ to $1$. As can be observed in Figs.~\ref{fig:compRatioLW_nmi} and \ref{fig:compRatioLW_ari}, moderate values of $\tau$ generally lead to better consensus clustering performance. When the sampling ratio $\tau$ goes from $0.8$ to $1$, the performance declines, which suggests that the use of random subspaces exhibits a positive influence when compared to using the full feature sets (by setting $\tau=1$). At the other extreme, when setting $\tau$ to very small values, e.g., in the range of [0.1, 0.3], the performance also declines, due to the fact that the subspaces generated by a very small sampling ratio may not well reflect the underlying structure of the dataset. Empirically, it is suggested that the sampling ratio $\tau$ be set in the range of $[0.4, 0.8]$, which strikes a balance between diversity and quality. In this paper, the sampling ratio $\tau=0.5$ is used in the experiments on all the benchmark datasets.

Third, we evaluate the performances of the proposed algorithms \emph{with} and \emph{without} the weighted clusters. Note that the performances of the proposed algorithms without weighted clusters are obtained by setting all cluster weights equal to one. As shown in Figs.~\ref{fig:compRatioLW_nmi} and \ref{fig:compRatioLW_ari}, in terms of both NMI and ARI, the proposed three algorithms with weighted clusters exhibit consistently better average performances (across the 30 benchmark datasets) than that without weighted clusters.

From the comparison results in Figs.~\ref{fig:compMetric}, \ref{fig:compRatioLW_nmi}, and \ref{fig:compRatioLW_ari}, we can have two main observations: 1) the performance of our approach benefits from the use of diversified metrics, random subspaces, and weighted clusters; 2) out of the three beneficial factors, the diversified metrics play the possibly most important role in the consensus clustering performance, with consideration to the clear improvement (w.r.t. both NMI and ARI) that they lead to.

\subsection{Execution Time}

In this section, we evaluate the efficiency of different ensemble clustering algorithms and report their execution times on the benchmark datasets in Table~\ref{table:compare_time}. The experiments are conducted in MATLAB R2016a 64-bit on a workstation with Intel i9-7940X CPU and 64 GB memory.

In general, larger dimensions and larger sample sizes lead to greater computational costs for the ensemble clustering algorithms.
As shown in Table~\ref{table:compare_time}, the proposed MDEC-HC, MDEC-SC, and MDEC-BG algorithms each consume less than 1 second of time on 14 out of the totally 18 cancer gene expression datasets. On the 12 image/speech datasets,  the proposed algorithms also show comparable time efficiency with the other ensemble clustering algorithms.

To summarize, as can be seen in Tables~\ref{table:compare_ce_nmi} to \ref{table:compare_time} and Figs.~\ref{fig:comp_nmi_Msize} to \ref{fig:compVaryM_all}, the proposed three ensemble clustering algorithms have shown considerable advantages in clustering effectiveness while exhibiting competitive time efficiency when compared with the state-of-the-art ensemble clustering algorithms.

\section{Conclusion}
\label{sec:conclude}

In this paper, we present a new ensemble clustering approach termed MDEC, which is capable of jointly exploiting large populations of diversified metrics, random subspaces, and weighted clusters in a unified ensemble clustering framework. Specifically, a large number of diversified metrics are generated by randomizing a scaled exponential similarity kernel. The diversified metrics are then coupled with the random subspaces to form a large set of metric-subspace pairs. Upon the similarity matrices derived from the metric-subspace pairs, the spectral clustering algorithm is performed to construct an ensemble of diversified base clusterings. With the base clusterings generated, an entropy-based cluster validity strategy is utilized to evaluate and weight the clusters with consideration to the distribution of the cluster labels in the entire ensemble. Finally, based on diversified metrics, random subspaces, and weighted clusters, three specific ensemble clustering algorithms are devised by incorporating three types of consensus functions. Extensive experiments are conducted on 30 real-world high-dimensional datasets (including 18 cancer gene expression datasets and 12 image/speech datasets), which have demonstrated the advantages of the proposed algorithms over the state-of-the-art ensemble clustering algorithms.

Note that in the practical application process of the proposed algorithms (as well as other ensemble clustering algorithms), two types of balance should be taken into consideration, i.e., the balance between diversity and quality (which involves enhancing different levels of diversity while maintaining the quality of the individuals) and the balance between effectiveness and efficiency (which involves the choice of base clusterers and the size of the ensembles).

In the future work, there are several possible research directions. First, starting from this work, more metric diversification strategies can be studied to seek more opportunities in metric spaces for ensemble clustering of high-dimensional data. Second, different consensus methods for exploiting multi-level diversity in ensembles can be investigated to deal with different types of complex data. Last but not least, the diversification-and-fusion strategy may be generalized to other unsupervised learning tasks (such as unsupervised feature selection and unsupervised outlier detection) to alleviate the difficulties in parameter selection, metric selection, and model selection.

\ifCLASSOPTIONcompsoc
  \section*{Acknowledgments}
\else
  \section*{Acknowledgment}
\fi

This project was supported by NSFC (61976097, 61876193 \& 61876104) and A*STAR-NTU-SUTD AI Partnership Grant (No. RGANS1905).

\begin{table*}[!t]\scriptsize
\centering
\caption{The execution times (in seconds) of different ensemble clustering algorithms
(with both ensemble generation and consensus function included).}
\label{table:compare_time}
\begin{tabular}{m{0.84cm}<{\centering}|m{0.82cm}<{\centering}m{0.82cm}<{\centering}m{0.82cm}<{\centering}m{0.82cm}<{\centering}m{0.82cm}<{\centering}m{0.82cm}<{\centering}m{0.82cm}<{\centering}m{0.82cm}<{\centering}m{0.9cm}<{\centering}m{1.3cm}<{\centering}m{1.3cm}<{\centering}m{1.3cm}<{\centering}}
\toprule
Dataset &SSCSPA     &SSHGPA     &SSMCLA     &KCC     &PTGP  &SEC    &ECC    &LWGP   &SPCE	&MDEC-HC   &MDEC-SC   &MDEC-BG\\
\midrule
\emph{DS-1}	&1.12	&1.51	&1.38	&0.87	&1.25	&0.80	&0.95	&1.65	&1.80	&1.00	&1.01	&1.07\\
\emph{DS-2}	&0.82	&1.02	&1.01	&0.44	&0.62	&0.42	&0.45	&1.38	&0.53	&0.75	&0.76	&0.77\\
\emph{DS-3}	&0.91	&1.04	&1.08	&0.45	&0.69	&0.42	&0.46	&1.10	&0.79	&0.77	&0.78	&0.80\\
\emph{DS-4}	&1.12	&1.31	&1.31	&0.64	&0.81	&0.61	&0.68	&1.24	&1.05	&0.87	&0.88	&0.90\\
\emph{DS-5}	&0.86	&1.02	&1.06	&0.52	&0.73	&0.49	&0.52	&1.14	&0.71	&0.78	&0.79	&0.80\\
\emph{DS-6}	&0.86	&1.08	&1.06	&0.54	&0.71	&0.50	&0.56	&1.44	&0.79	&0.78	&0.78	&0.80\\
\emph{DS-7}	&1.13	&1.00	&0.99	&0.46	&0.64	&0.43	&0.48	&1.04	&0.58	&0.73	&0.74	&0.75\\
\emph{DS-8}	&0.80	&0.97	&0.97	&0.46	&0.61	&0.43	&0.48	&1.01	&0.56	&0.71	&0.72	&0.72\\
\emph{DS-9}	&1.06	&1.52	&1.28	&0.79	&1.04	&0.68	&0.90	&1.47	&1.72	&0.90	&0.91	&0.95\\
\emph{DS-10}	&0.80	&0.93	&0.96	&0.45	&0.61	&0.43	&0.45	&1.01	&0.51	&0.72	&0.73	&0.74\\
\emph{DS-11}	&1.68	&1.78	&1.89	&1.31	&1.74	&1.23	&1.37	&2.15	&4.01	&1.14	&1.15	&1.21\\
\emph{DS-12}	&1.52	&1.81	&1.75	&1.34	&1.68	&1.20	&1.50	&2.10	&3.94	&1.15	&1.17	&1.23\\
\emph{DS-13}	&0.85	&1.09	&1.03	&0.49	&0.67	&0.47	&0.50	&1.07	&0.63	&0.74	&0.74	&0.76\\
\emph{DS-14}	&0.86	&1.09	&1.03	&0.52	&0.69	&0.49	&0.54	&1.09	&0.64	&0.74	&0.75	&0.76\\
\emph{DS-15}	&0.89	&1.12	&1.09	&0.52	&0.71	&0.49	&0.54	&1.14	&0.76	&1.16	&1.17	&1.19\\
\emph{DS-16}	&1.07	&1.48	&1.40	&0.91	&1.75	&0.75	&1.06	&2.26	&1.97	&0.87	&0.89	&0.93\\
\emph{DS-17}	&0.82	&0.98	&0.99	&0.44	&0.58	&0.41	&0.45	&0.99	&0.55	&0.73	&0.74	&0.75\\
\emph{DS-18}	&1.02	&1.33	&1.22	&0.66	&0.89	&0.62	&0.70	&1.31	&1.21	&0.83	&0.84	&0.87\\
\midrule
\emph{DS-19}	&41.32	&90.94	&26.43	&61.71	&156.60	&43.51	&87.18	&59.94	&4032.33	&137.40	&144.16	&139.01\\
\emph{DS-20}	&3.43	&4.22	&3.09	&5.32	&4.89	&4.15	&6.58	&7.41	&114.84	&5.44	&5.71	&5.84\\
\emph{DS-21}	&3.43	&3.58	&2.91	&2.82	&2.72	&1.48	&4.26	&5.39	&178.47	&6.89	&7.18	&7.37\\
\emph{DS-22}	&2.98	&43.10	&3.37	&3.52	&3.53	&2.04	&5.78	&5.72	&106.74	&6.09	&6.56	&6.48\\
\emph{DS-23}	&9.55	&16.17	&6.84	&17.40	&19.09	&13.82	&20.82	&18.93	&354.36	&15.66	&16.92	&16.32\\
\emph{DS-24}	&3.42	&5.85	&3.54	&3.78	&4.54	&2.37	&5.80	&6.33	&90.75	&7.23	&7.47	&7.64\\
\emph{DS-25}	&5.92	&7.04	&6.18	&13.46	&13.43	&13.11	&13.90	&15.21	&29.72	&4.72	&4.79	&4.89\\
\emph{DS-26}	&5.58	&5.36	&4.33	&6.80	&7.66	&4.73	&8.87	&9.37	&140.78	&9.49	&9.82	&10.06\\
\emph{DS-27}	&44.27	&46.46	&44.12	&204.92	&204.92	&203.34	&206.45	&206.91	&299.38	&31.26	&31.51	&31.64\\
\emph{DS-28}	&33.05	&31.17	&18.75	&43.00	&69.10	&33.34	&53.65	&44.77	&1115.70	&66.18	&68.29	&67.67\\
\emph{DS-29}	&117.37	&95.22	&97.91	&450.27	&602.59	&441.79	&464.34	&459.80	&3590.84	&238.26	&240.38	&239.67\\
\emph{DS-30}	&74.94	&79.57	&37.49	&65.89	&590.31	&43.67	&109.65	&70.75	&OM	&302.52	&312.08	&305.36\\
\bottomrule
\end{tabular}
\end{table*}

\ifCLASSOPTIONcaptionsoff
  \newpage
\fi

\bibliographystyle{IEEEtran}
\bibliography{tcyb_2019}

\begin{thebibliography}{10}
\providecommand{\url}[1]{#1}
\csname url@samestyle\endcsname
\providecommand{\newblock}{\relax}
\providecommand{\bibinfo}[2]{#2}
\providecommand{\BIBentrySTDinterwordspacing}{\spaceskip=0pt\relax}
\providecommand{\BIBentryALTinterwordstretchfactor}{4}
\providecommand{\BIBentryALTinterwordspacing}{\spaceskip=\fontdimen2\font plus
\BIBentryALTinterwordstretchfactor\fontdimen3\font minus
  \fontdimen4\font\relax}
\providecommand{\BIBforeignlanguage}[2]{{%
\expandafter\ifx\csname l@#1\endcsname\relax
\typeout{** WARNING: IEEEtran.bst: No hyphenation pattern has been}%
\typeout{** loaded for the language `#1'. Using the pattern for}%
\typeout{** the default language instead.}%
\else
\language=\csname l@#1\endcsname
\fi
#2}}
\providecommand{\BIBdecl}{\relax}
\BIBdecl

\bibitem{Li_WCC08}
T.~Li and C.~Ding, ``Weighted consensus clustering,'' in \emph{Proc. of SIAM
  International Conference on Data Mining (SDM)}, 2008, pp. 798--809.

\bibitem{iam_on11_linkbased}
N.~Iam-On, T.~Boongoen, S.~Garrett, and C.~Price, ``A link-based approach to
  the cluster ensemble problem,'' \emph{IEEE Transactions on Pattern Analysis
  and Machine Intelligence}, vol.~33, no.~12, pp. 2396--2409, 2011.

\bibitem{wang11_tsmcb}
T.~Wang, ``{CA-T}ree: A hierarchical structure for efficient and scalable
  coassociation-based cluster ensembles,'' \emph{IEEE Transactions on Systems,
  Man, and Cybernetics, Part B: Cybernetics}, vol.~41, no.~3, pp. 686--698,
  2011.

\bibitem{li12_nips}
N.~Li and L.~J. Latecki, ``Clustering aggregation as maximum-weight independent
  set,'' in \emph{Advances in Neural Information Processing Systems (NIPS)},
  2012, pp. 782--790.

\bibitem{Zheng14_TKDD}
L.~Zheng, T.~Li, and C.~Ding, ``A framework for hierarchical ensemble
  clustering,'' \emph{ACM Transactions on Knowledge Discovery from Data},
  vol.~9, no.~2, pp. 9:1--9:23, 2014.

\bibitem{Jing15_pr}
L.~Jing, K.~Tian, and J.~Z. Huang, ``Stratified feature sampling method for
  ensemble clustering of high dimensional data,'' \emph{Pattern Recognition},
  vol.~48, no.~11, pp. 3688--3702, 2015.

\bibitem{wu15_TKDE}
J.~Wu, H.~Liu, H.~Xiong, J.~Cao, and J.~Chen, ``K-means-based consensus
  clustering: A unified view,'' \emph{IEEE Transactions on Knowledge and Data
  Engineering}, vol.~27, no.~1, pp. 155--169, 2015.

\bibitem{Huang16_TKDE}
D.~Huang, J.-H. Lai, and C.-D. Wang, ``Robust ensemble clustering using
  probability trajectories,'' \emph{IEEE Transactions on Knowledge and Data
  Engineering}, vol.~28, no.~5, pp. 1312--1326, 2016.

\bibitem{huang15_ecfg}
D.~Huang, J.~Lai, and C.-D. Wang, ``Ensemble clustering using factor graph,''
  \emph{Pattern Recognition}, vol.~50, pp. 131--142, 2016.

\bibitem{Liu2017_DMKD}
H.~Liu, M.~Shao, S.~Li, and Y.~Fu, ``Infinite ensemble clustering,'' \emph{Data
  Mining and Knowledge Discovery}, vol.~32, pp. 385--416, 2018.

\bibitem{liu17_tkde}
H.~Liu, J.~Wu, T.~Liu, D.~Tao, and Y.~Fu, ``Spectral ensemble clustering via
  weighted k-means: Theoretical and practical evidence,'' \emph{IEEE
  Transactions on Knowledge and Data Engineering}, vol.~29, no.~5, pp.
  1129--1143, 2017.

\bibitem{liu17_bioinformatics}
H.~Liu, R.~Zhao, H.~Fang, F.~Cheng, Y.~Fu, and Y.-Y. Liu, ``Entropy-based
  consensus clustering for patient stratification,'' \emph{Bioinformatics},
  vol.~33, no.~17, pp. 2691--2698, 2017.

\bibitem{yu18_tkde}
Z.~Yu, P.~{Luo}, J.~{Liu}, H.~{Wong}, J.~{You}, G.~{Han}, and J.~{Zhang},
  ``Semi-supervised ensemble clustering based on selected constraint
  projection,'' \emph{IEEE Transactions on Knowledge and Data Engineering},
  vol.~30, no.~12, pp. 2394--2407, 2018.

\bibitem{shi18_tcyb}
Y.~{Shi}, Z.~{Yu}, C.~L.~P. {Chen}, J.~{You}, H.~{Wong}, Y.~{Wang}, and
  J.~{Zhang}, ``Transfer clustering ensemble selection,'' \emph{IEEE
  Transactions on Cybernetics}, vol.~50, no.~6, pp. 2872--2885, 2020.

\bibitem{huang17_tcyb}
D.~Huang, C.~D. Wang, and J.~H. Lai, ``Locally weighted ensemble clustering,''
  \emph{IEEE Transactions on Cybernetics}, vol.~48, no.~5, pp. 1460--1473,
  2018.

\bibitem{huang18_tsmcs}
D.~Huang, C.-D. {Wang}, H.~{Peng}, J.-H. {Lai}, and C.-K. {Kwoh}, ``Enhanced
  ensemble clustering via fast propagation of cluster-wise similarities,''
  \emph{IEEE Transactions on Systems, Man, and Cybernetics: Systems}, 2021.

\bibitem{huang19_tkde}
D.~Huang, C.-D. {Wang}, J.-S. {Wu}, J.-H. {Lai}, and C.-K. {Kwoh},
  ``Ultra-scalable spectral clustering and ensemble clustering,'' \emph{IEEE
  Transactions on Knowledge and Data Engineering}, vol.~32, no.~6, pp.
  1212--1226, 2020.

\bibitem{bai18_tfs}
L.~Bai, J.~{Liang}, and Y.~{Guo}, ``An ensemble clusterer of multiple fuzzy $k$
  -means clusterings to recognize arbitrarily shaped clusters,'' \emph{IEEE
  Transactions on Fuzzy Systems}, vol.~26, no.~6, pp. 3524--3533, 2018.

\bibitem{bai19_tkde}
L.~Bai, J.~{Liang}, H.~{Du}, and Y.~{Guo}, ``An information-theoretical
  framework for cluster ensemble,'' \emph{IEEE Transactions on Knowledge and
  Data Engineering}, vol.~31, no.~8, pp. 1464--1477, 2019.

\bibitem{tao19_tnnls}
Z.~Tao, H.~{Liu}, S.~{Li}, Z.~{Ding}, and Y.~{Fu}, ``Marginalized multiview
  ensemble clustering,'' \emph{IEEE Transactions on Neural Networks and
  Learning Systems}, 2019.

\bibitem{yu19_tkde}
X.~Yu, G.~{Yu}, J.~{Wang}, and C.~{Domeniconi}, ``Co-clustering ensembles based
  on multiple relevance measures,'' \emph{IEEE Transactions on Knowledge and
  Data Engineering, in press}, 2019.

\bibitem{li19_ai}
F.~Li, Y.~Qian, J.~Wang, C.~Dang, and L.~Jing, ``Clustering ensemble based on
  sample's stability,'' \emph{Artificial Intelligence}, vol. 273, pp. 37--55,
  2019.

\bibitem{Yu07_bioinformatics}
Z.~Yu, H.-S. Wong, and H.~Wang, ``Graph-based consensus clustering for class
  discovery from gene expression data,'' \emph{Bioinformatics}, vol.~23,
  no.~21, pp. 2888--2896, 2007.

\bibitem{yu15_tkde}
Z.~Yu, L.~Li, J.~Liu, J.~Zhang, and G.~Han, ``Adaptive noise immune cluster
  ensemble using affinity propagation,'' \emph{IEEE Transactions on Knowledge
  and Data Engineering}, vol.~27, no.~12, pp. 3176--3189, 2015.

\bibitem{Yu16_tkde_incremental}
Z.~Yu, P.~Luo, J.~You, H.~S. Wong, H.~Leung, S.~Wu, J.~Zhang, and G.~Han,
  ``Incremental semi-supervised clustering ensemble for high dimensional data
  clustering,'' \emph{IEEE Transactions on Knowledge and Data Engineering},
  vol.~28, no.~3, pp. 701--714, 2016.

\bibitem{Yu17_tkde}
Z.~Yu, Z.~Kuang, J.~Liu, H.~Chen, J.~Zhang, J.~You, H.~S. Wong, and G.~Han,
  ``Adaptive ensembling of semi-supervised clustering solutions,'' \emph{IEEE
  Transactions on Knowledge and Data Engineering}, vol.~29, no.~8, pp.
  1577--1590, 2017.

\bibitem{fern2003random}
X.~Z. Fern and C.~E. Brodley, ``Random projection for high dimensional data
  clustering: A cluster ensemble approach,'' in \emph{Proc. of International
  Conference on Machine Learning (ICML)}, 2003, pp. 186--193.

\bibitem{Lee14_TVCG}
J.~H. Lee, K.~T. McDonnell, A.~Zelenyuk, D.~Imre, and K.~Mueller, ``A
  structure-based distance metric for high-dimensional space exploration with
  multidimensional scaling,'' \emph{IEEE Transactions on Visualization and
  Computer Graphics}, vol.~20, no.~3, pp. 351--364, 2014.

\bibitem{Hsu09}
C.~M. Hsu and M.~S. Chen, ``On the design and applicability of distance
  functions in high-dimensional data space,'' \emph{IEEE Transactions on
  Knowledge and Data Engineering}, vol.~21, no.~4, pp. 523--536, 2009.

\bibitem{Fred05_EAC}
A.~L.~N. Fred and A.~K. Jain, ``Combining multiple clusterings using evidence
  accumulation,'' \emph{IEEE Transactions on Pattern Analysis and Machine
  Intelligence}, vol.~27, no.~6, pp. 835--850, 2005.

\bibitem{wu13_ijcai}
J.~Wu, H.~Liu, H.~Xiong, and J.~Cao, ``A theoretic framework of k-means-based
  consensus clustering,'' in \emph{Proc. of International Joint Conference on
  Artificial Intelligence}, 2013.

\bibitem{Zhong15_pr}
C.~Zhong, X.~Yue, Z.~Zhang, and J.~Lei, ``A clustering ensemble:
  Two-level-refined co-association matrix with path-based transformation,''
  \emph{Pattern Recognition}, vol.~48, no.~8, pp. 2699--2709, 2015.

\bibitem{huang14_weac}
D.~Huang, J.-H. Lai, and C.-D. Wang, ``Combining multiple clusterings via crowd
  agreement estimation and multi-granularity link analysis,''
  \emph{Neurocomputing}, vol. 170, pp. 240--250, 2015.

\bibitem{Fan17_ijcai}
Y.~Fan, N.~Li, C.~Li, Z.~Ma, L.~J. Latecki, and K.~Su, ``Restart and random
  walk in local search for maximum vertex weight cliques with evaluations in
  clustering aggregation,'' in \emph{Proc. of International Joint Conference on
  Artificial Intelligence (IJCAI)}, 2017, pp. 622--630.

\bibitem{Yousefnezhad17_tcyb}
M.~Yousefnezhad, S.~J. Huang, and D.~Zhang, ``Wo{CE}: A framework for
  clustering ensemble by exploiting the wisdom of crowds theory,'' \emph{IEEE
  Transactions on Cybernetics}, vol.~48, no.~2, pp. 486--499, 2018.

\bibitem{Huang16_neucom}
D.~Huang, J.-H. Lai, C.-D. Wang, and P.~C. Yuen, ``Ensembling
  over-segmentations: From weak evidence to strong segmentation,''
  \emph{Neurocomputing}, vol. 207, pp. 416--427, 2016.

\bibitem{wang09_pr}
X.~Wang, C.~Yang, and J.~Zhou, ``Clustering aggregation by probability
  accumulation,'' \emph{Pattern Recognition}, vol.~42, no.~5, pp. 668--675,
  2009.

\bibitem{yi_icdm12}
J.~Yi, T.~Yang, R.~Jin, and A.~K. Jain, ``Robust ensemble clustering by matrix
  completion,'' in \emph{Proc. of IEEE International Conference on Data Mining
  (ICDM)}, 2012.

\bibitem{strehl02}
A.~Strehl and J.~Ghosh, ``Cluster ensembles: A knowledge reuse framework for
  combining multiple partitions,'' \emph{Journal of Machine Learning Research},
  vol.~3, pp. 583--617, 2003.

\bibitem{fern04_bipartite}
X.~Z. Fern and C.~E. Brodley, ``Solving cluster ensemble problems by bipartite
  graph partitioning,'' in \emph{Proc. of International Conference on Machine
  Learning (ICML)}, 2004.

\bibitem{topchy05}
A.~Topchy, A.~K. Jain, and W.~Punch, ``Clustering ensembles: models of
  consensus and weak partitions,'' \emph{IEEE Transactions on Pattern Analysis
  and Machine Intelligence}, vol.~27, no.~12, pp. 1866--1881, 2005.

\bibitem{franek13_pr}
L.~Franek and X.~Jiang, ``Ensemble clustering by means of clustering embedding
  in vector spaces,'' \emph{Pattern Recognition}, vol.~47, no.~2, pp. 833--842,
  2014.

\bibitem{jain10_survey}
A.~K. Jain, ``Data clustering: 50 years beyond $k$-means,'' \emph{Pattern
  Recognition Letters}, vol.~31, no.~8, pp. 651--666, 2010.

\bibitem{karypis98_METIS}
G.~Karypis and V.~Kumar, ``A fast and high quality multilevel scheme for
  partitioning irregular graphs,'' \emph{SIAM Journal on Scientific Computing},
  vol.~20, no.~1, pp. 359--392, 1998.

\bibitem{Weiszfeld09}
E.~Weiszfeld and F.~Plastria, ``On the point for which the sum of the distances
  to n given points is minimum,'' \emph{Annals of Operations Research}, vol.
  167, no.~1, pp. 7--41, 2009.

\bibitem{Kschischang_FG_SPA:01}
F.~R. Kschischang, B.~J. Frey, and H.-A. Loeliger, ``Factor graphs and the
  sum-product algorithm,'' \emph{IEEE Transactions on Information Theory},
  vol.~47, no.~2, pp. 498--519, 2001.

\bibitem{chu20_tkde}
J.~{Chu}, H.~{Wang}, J.~{Liu}, Z.~{Gong}, and T.~{Li}, ``Unsupervised feature
  learning architecture with multi-clustering integration {RBM},'' \emph{IEEE
  Transactions on Knowledge and Data Engineering}, 2020.

\bibitem{Wang13_pcyuen}
Q.~Wang, P.~C. Yuen, and G.~Feng, ``Semi-supervised metric learning via
  topology preserving multiple semi-supervised assumptions,'' \emph{Pattern
  Recognition}, vol.~46, no.~9, pp. 2576--2587, 2013.

\bibitem{xiong16_tkdd}
F.~Xiong, M.~Kam, L.~Hrebien, B.~Wang, and Y.~Qi, ``Kernelized
  information-theoretic metric learning for cancer diagnosis using
  high-dimensional molecular profiling data,'' \emph{ACM Transactions on
  Knowledge Discovery from Data}, vol.~10, no.~4, 2016.

\bibitem{li18_ijcv}
J.~Li, A.~J. Ma, and P.~C. Yuen, ``Semi-supervised region metric learning for
  person re-identification,'' \emph{International Journal of Computer Vision},
  vol. 126, pp. 855--874, 2018.

\bibitem{liu19_tpami}
X.~Liu, L.~{Wang}, X.~{Zhu}, M.~{Li}, E.~{Zhu}, T.~{Liu}, L.~{Liu}, Y.~{Dou},
  and J.~{Yin}, ``Absent multiple kernel learning algorithms,'' \emph{IEEE
  Transactions on Pattern Analysis and Machine Intelligence, in press}, 2019.

\bibitem{chen19_tcyb}
X.~Chen, W.~{Sun}, B.~{Wang}, Z.~{Li}, X.~{Wang}, and Y.~{Ye}, ``Spectral
  clustering of customer transaction data with a two-level subspace weighting
  method,'' \emph{IEEE Transactions on Cybernetics}, vol.~49, no.~9, pp.
  3230--3241, 2019.

\bibitem{han19_nips}
H.~Liu, Z.~Han, Y.-S. Liu, and M.~Gu, ``Fast low-rank metric learning for
  large-scale and high-dimensional data,'' in \emph{Advances in Neural
  Information Processing Systems (NeurIPS)}, 2019, pp. 819--829.

\bibitem{wang14_nmeth}
B.~Wang, A.~M. Mezlini, F.~Demir, M.~Fiume, Z.~Tu, M.~Brudno, B.~Haibe-Kains,
  and A.~Goldenberg, ``Similarity network fusion for aggregating data types on
  a genomic scale,'' \emph{Nature Methods}, vol.~11, pp. 333--337, 2014.

\bibitem{vonLuxburg2007}
U.~von Luxburg, ``A tutorial on spectral clustering,'' \emph{Statistics and
  Computing}, vol.~17, no.~4, pp. 395--416, 2007.

\bibitem{CVPR12_Li}
Z.~Li, X.-M. Wu, and S.-F. Chang, ``Segmentation using superpixels: A bipartite
  graph partitioning approach,'' in \emph{Proc. of IEEE Conference on Computer
  Vision and Pattern Recognition (CVPR)}, 2012.

\bibitem{deSouto08_gds}
M.~C. de~Souto, I.~G. Costa, D.~S. de~Araujo, T.~B. Ludermir, and A.~Schliep,
  ``Clustering cancer gene expression data: A comparative study,'' \emph{BMC
  bioinformatics}, vol.~9, no.~1, p. 497, 2008.

\bibitem{Bache+Lichman:2013}
\BIBentryALTinterwordspacing
K.~Bache and M.~Lichman, ``{UCI} machine learning repository,'' 2017. [Online].
  Available: \url{http://archive.ics.uci.edu/ml}
\BIBentrySTDinterwordspacing

\bibitem{nene96_coil20}
S.~A. Nene, S.~K. Nayar, H.~Murase \emph{et~al.}, ``Columbia object image
  library ({COIL}-20),'' \emph{Technical report CUCS-005-96}, 1996.

\bibitem{nie19_tcyb}
F.~Nie, Z.~{Wang}, R.~{Wang}, and X.~{Li}, ``Submanifold-preserving
  discriminant analysis with an auto-optimized graph,'' \emph{IEEE Transactions
  on Cybernetics, in press}, 2019.

\bibitem{SamRoweisHomepage}
S.~Roweis, \url{http://www.cs.nyu.edu/\%7eroweis/data.html}.

\bibitem{graham98_UMist}
D.~B. Graham and N.~M. Allinson, ``Characterising virtual eigensignatures for
  general purpose face recognition,'' in \emph{Face Recognition}.\hskip 1em
  plus 0.5em minus 0.4em\relax Springer, 1998, pp. 446--456.

\bibitem{nilsback06_flower17}
M.-E. Nilsback and A.~Zisserman, ``A visual vocabulary for flower
  classification,'' in \emph{Proc. of IEEE Conference on Computer Vision and
  Pattern Recognition (CVPR)}, vol.~2, 2006, pp. 1447--1454.

\bibitem{vinh2010_ARI}
N.~X. Vinh, J.~Epps, and J.~Bailey, ``Information theoretic measures for
  clusterings comparison: Variants, properties, normalization and correction
  for chance,'' \emph{Journal of Machine Learning Research}, vol.~11, no.~11,
  pp. 2837--2854, 2010.

\bibitem{zhou20_spce}
P.~Zhou, L.~Du, X.~Liu, Y.-D. Shen, M.~Fan, and X.~Li, ``Self-paced clustering
  ensemble,'' \emph{IEEE Transactions on Neural Networks and Learning Systems,
  in press}, 2020.

\end{thebibliography}

\begin{IEEEbiography}[{\includegraphics[width=1in,height=1.25in,clip,keepaspectratio]{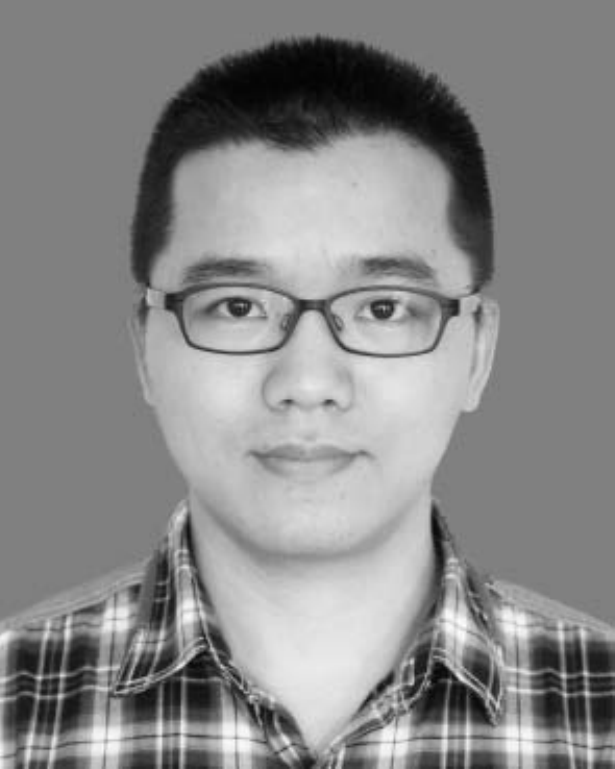}}]{Dong Huang}
received the B.S. degree in computer science from South China University of Technology, Guangzhou, China, in 2009, and the M.Sc. and Ph.D. degrees in computer science from Sun Yat-sen University, Guangzhou, in 2011 and 2015, respectively. He joined South China Agricultural University, Guangzhou, in 2015, where he is currently working as an Associate Professor and Deputy Head in the Department of Computer Science. From 2017 to 2018, he was a visiting fellow with the School of Computer Science and Engineering, Nanyang Technological University, Singapore. His current research interests include data mining and machine learning, and more specifically focus on ensemble clustering, multi-view clustering, and large-scale clustering. He has published more than 50 papers in refereed journals and conferences, including IEEE TKDE, IEEE TCYB, IEEE TSMC-S, ACM TKDD, Pattern Recognition, KBS, AAAI, and ICDM. He is a member of the IEEE.
\end{IEEEbiography}

\begin{IEEEbiography}[{\includegraphics[width=1in,height=1.25in,clip,keepaspectratio]{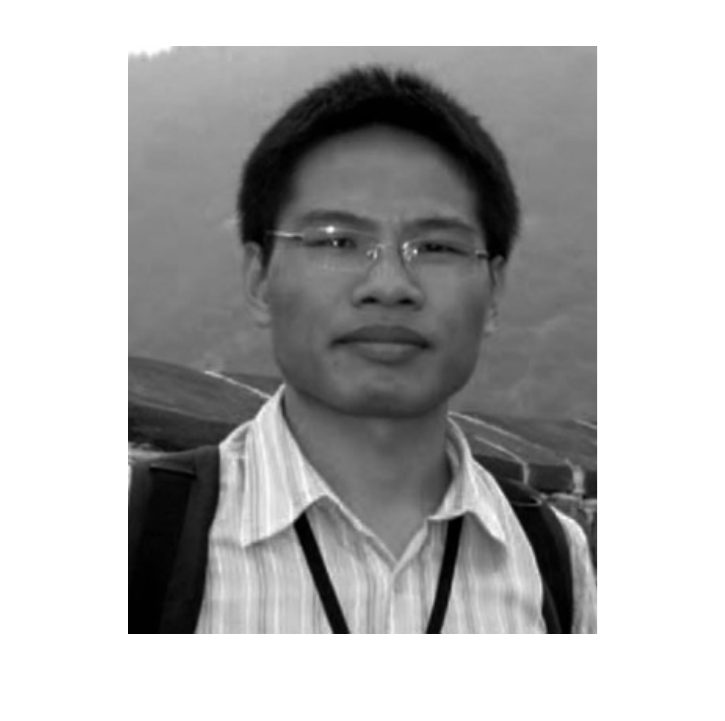}}]{Chang-Dong Wang}
received the B.S. degree in applied mathematics in 2008, the M.Sc. degree in computer science in 2010, and the Ph.D. degree in computer science in 2013, all from Sun Yat-sen University, Guangzhou, China. He was a visiting student at the University of Illinois at Chicago from January 2012 to November 2012. He is currently an Associate Professor with the School of Data and Computer Science, Sun Yat-sen University, Guangzhou, China. His current research interests include machine learning and data mining. He has published more than 100 scientific papers in international journals and conferences such as IEEE TPAMI, IEEE TKDE, IEEE TCYB, ACM TKDD, Pattern Recognition, KAIS, IJCAI, AAAI, KDD, ICDM and SDM. His ICDM 2010 paper won the Honorable Mention for Best Research Paper Awards. He was awarded 2015 Chinese Association for Artificial Intelligence (CAAI) Outstanding Dissertation. He is a member of the IEEE.
\end{IEEEbiography}

\begin{IEEEbiography}[{\includegraphics[width=1in,height=1.25in,clip,keepaspectratio]{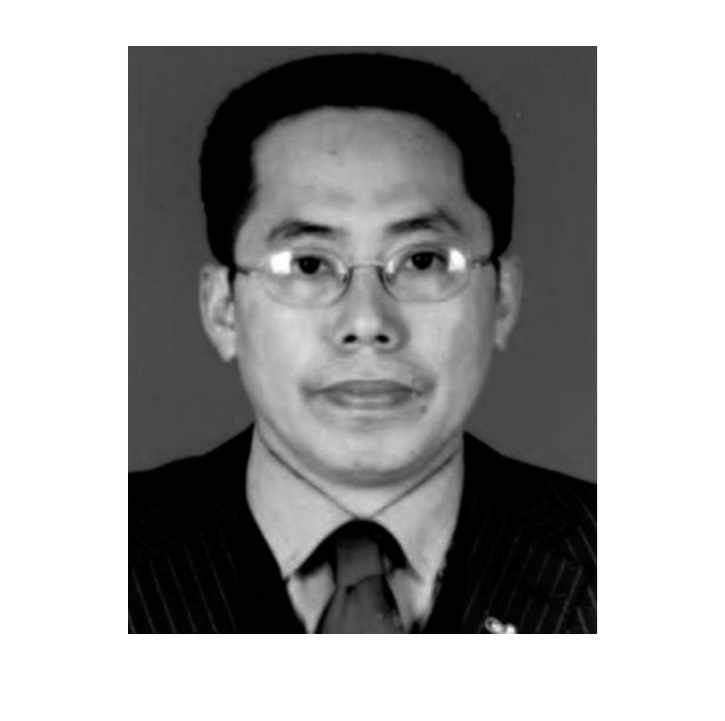}}]{Jian-Huang Lai}
received the M.Sc. degree in applied mathematics in 1989 and the Ph.D. degree in mathematics in 1999 from Sun Yat-sen University, China. He joined Sun Yat-sen University in 1989 as an Assistant Professor, where he is currently a Professor with the School of Data and Computer Science. His current research interests include the areas of digital image processing, pattern recognition, multimedia communication, wavelet and its applications. He has published more than 200 scientific papers in the international journals and conferences on image processing and pattern recognition, such as IEEE TPAMI, IEEE TKDE, IEEE TNN, IEEE TIP, Pattern Recognition, ICCV, CVPR, IJCAI, ICDM and SDM. Prof. Lai serves as a Standing Member of the Image and Graphics Association of China, and also serves as a Standing Director of the Image and Graphics Association of Guangdong. He is a senior member of the IEEE.
\end{IEEEbiography}

\begin{IEEEbiography}[{\includegraphics[width=1in,height=1.25in,clip,keepaspectratio]{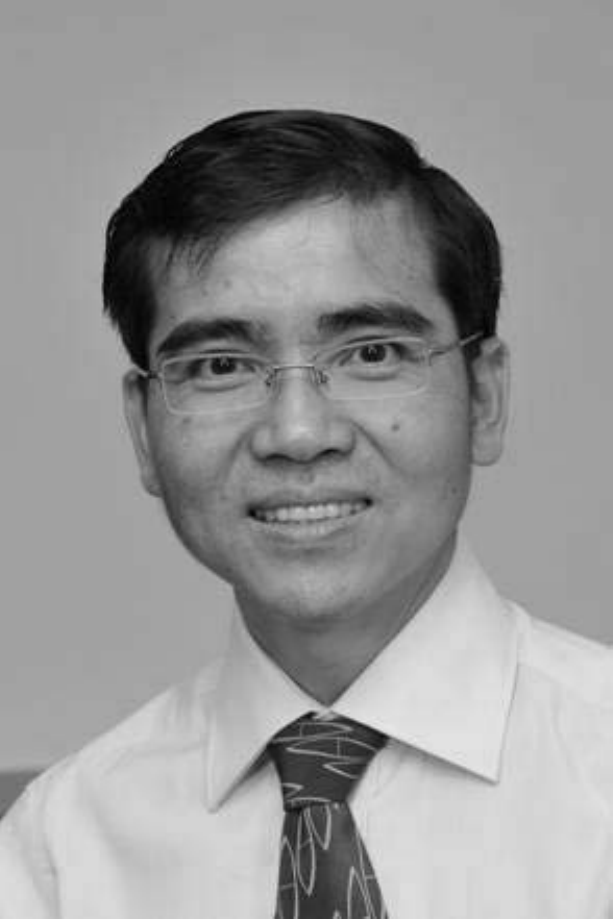}}]{Chee-Keong Kwoh} received the bachelor's
degree in electrical engineering (first class) and the master's degree in industrial system engineering from the National University of Singapore in 1987 and 1991, respectively. He received the PhD degree from the Imperial College of Science, Technology and Medicine, University of London, in 1995. He has been with the School of Computer Science and Engineering, Nanyang Technological University (NTU) since 1993. He is the programme director of the MSc in Bioinformatics programme at NTU. His research interests include data mining, soft computing and graph-based inference; applications areas include bioinformatics and biomedical engineering. He has done significant research work in his research areas and has published many quality international conferences and journal papers. He is an editorial board member of the International Journal of Data Mining and Bioinformatics; the Scientific World Journal; Network Modeling and Analysis in Health Informatics and Bioinformatics; Theoretical Biology Insights; and Bioinformation. He has been a guest editor for many journals such as Journal of Mechanics in Medicine and Biology, the International Journal on Biomedical and Pharmaceutical Engineering
and others. He is a member of the Association for Medical and Bio-Informatics, Imperial College Alumni Association of Singapore. He has provided many services to professional bodies in Singapore and was conferred the Public Service Medal by the President of Singapore in 2008. He is a senior member of the IEEE.
\end{IEEEbiography}

\end{document}